\documentclass[final,5p,times,authoryear,monochrome]{elsarticle}

\usepackage[utf8]{inputenc}
\usepackage[english]{babel}
\usepackage{pdfpages}
\usepackage{hyperref}
\usepackage{graphicx}
\usepackage{float}
\usepackage{amsmath,amsthm}
\usepackage{amssymb}

\usepackage {tikz}
\usetikzlibrary {positioning}
\usetikzlibrary{shapes.misc, positioning}
\usetikzlibrary{arrows.meta}
\definecolor {processblue}{cmyk}{0.96,0,0,0}
\usepackage{subcaption}
\usepackage[export]{adjustbox}

\usepackage{xcolor}

\usepackage{cclicenses}  
\newcommand\blfootnote[1]{%
  \begingroup
  \renewcommand\thefootnote{}\footnote{#1}%
  \addtocounter{footnote}{-1}%
  \endgroup
}

\journal{Computers \& Operations Research}

\begin{document}

\begin{frontmatter}

\title{Evolving test instances of the
	Hamiltonian completion problem}

\author{Thibault Lechien\corref{cor1}\fnref{kul}}
\ead{thibault.lechien@student.kuleuven.be}

\author{Jorik Jooken\fnref{kul}}
\ead{jorik.jooken@kuleuven.be}

\author{Patrick De Causmaecker\fnref{kul}}
\ead{patrick.decausmaecker@kuleuven.be}

\affiliation[kul]{organization={Department of Computer Science, CODeS, KU Leuven Kulak},%Department and Organization
	addressline={Etienne Sabbelaan 53}, 
	city={Kortrijk},
	postcode={8500}, 
	country={Belgium}}

\cortext[cor1]{Corresponding author}

	\begin{abstract}
	Predicting and comparing algorithm performance on graph instances is challenging for multiple reasons. First, there is not always a standard set of instances to benchmark performance. Second, using existing graph generators results in a restricted spectrum of difficulty and the resulting graphs are not always diverse enough to draw sound conclusions. That is why recent work proposes a new methodology to generate a diverse set of instances by using evolutionary algorithms. We can then analyze the resulting graphs and get key insights into which attributes are most related to algorithm performance. We can also fill observed gaps in the instance space in order to generate graphs with previously unseen combinations of features. We apply this methodology to the instance space of the Hamiltonian completion problem using two different solvers, namely the Concorde TSP Solver and a multi-start local search algorithm.
	\end{abstract}
	
	
	%%Research highlights (include the commented lines if more than 85 characters allowed, as they are less than 20 words)
%	\begin{highlights}
%		\item Applied a novel methodology to generate more diverse and challenging test instances 
%		%than standard graph generators
%		\item Visualized the instance space, providing insight into each algorithm's performance
%		%Visualized the instance space which provides a better understanding to each algorithm's strengths and weaknesses
%		\item Showed that reducing dimensions retains key characteristics
%		% including the ability to predict algorithm performance
%	\end{highlights}
	
\end{frontmatter}

	\section{Introduction}
	\blfootnote{This work is licensed under a Creative Commons Attribution-NonCommercial-NoDerivatives 4.0 International License \cc \byncnd}\blfootnote{DOI: \url{https://doi.org/10.1016/j.cor.2022.106019}}The selection of test instances is crucial when evaluating algorithms. Predicting performance, gaining a deeper understanding of algorithm strengths and weaknesses, and helping design better algorithms all depend on those instances. Selecting or generating a diverse set of instances can be quite challenging however. A commonly used approach is to randomly generate test instances, but these often lack diversity or do not resemble real-world instances, as argued by \cite{HOOKER}. He also raised concerns about another approach, namely using well established benchmark instances (if these exist for the problem). This could play a part in the over-tuning of algorithms, which has a considerable impact on the applicability of their results.

	Instead of these methods, we use an evolutionary algorithm to generate instances that are more diverse and challenging for each solver. By evolving and examining instances that are easy or hard to solve for certain algorithms, we aim to get a better understanding of which features correlate with instance hardness. We do this by trying to maximize the difference in runtime. This provides a more detailed look into the respective strengths and weaknesses of existing algorithms. To our knowledge, this methodology was first demonstrated by \cite{Evolvinginstances} on three important domains of combinatorial optimization, namely boolean satisfiability \citep{sat}, binary constraint satisfaction \citep{constraint} and the traveling salesman problem \citep{tsp}. 
	After utilizing the evolutionary algorithm, we visualize the instance space by projecting the high-dimensional feature space down to two dimensions. This shows the similarities and differences between instances and enables us to identify gaps in the space, where there could be instances with previously unseen combinations of features. It also allows the performance of algorithms to be viewed and predicted across the space. This approach was demonstrated in \cite{SMITHMILES2015102} and \cite{EvolvedInstancesTSP}, using graph coloring and the traveling salesman problem as a case study respectively.

	We apply this methodology to the Hamiltonian completion problem (HCP) for undirected graphs \citep{hcp}. The objective for this optimization problem is to add as few edges as possible to a given undirected graph in order to obtain a Hamiltonian graph.
	A graph is Hamiltonian when it has at least one Hamiltonian cycle, which is a cycle that contains each node of the graph exactly once. HCP can for example be used to solve a special case of the traveling salesman problem \citep{TSPapplication} or to assign frequencies to transmitters \citep{transmitters}.
	Recently, a matheuristic was developed by \cite{jooken2019multistart}, which attempts to solve this problem heuristically by using a multi-start local search algorithm. HCP can also be solved by first converting it to a traveling salesman problem instance (TSP) and then using a TSP solver such as Concorde \citep{Concorde} or Lin-Kernighan-Helsgaun \citep{lkh} to get an exact or heuristic solution respectively. 
	In this paper we will compare the multi-start local search algorithm (\textit{MSLS}) \citep{jooken2019multistart} with the Concorde TSP solver (\textit{Concorde}).
	
	The main contributions of our work are summarized as follows.
	\textcolor{cyan}{We have evolved more challenging instances than standard graph generators, which allows us to find instances that MSLS solves faster than Concorde and vice versa. We also show that MSLS does not outperform Concorde in many cases, which is in contrast to the conclusions drawn in \cite{jooken2019multistart}. 
	One can only draw reasonable conclusions about the performance of an algorithm when testing on a broad and diverse set of instances, which is demonstrated in this paper. 
	The previous instances did not lead to a large difference in runtime between both solvers, whereas our evolved instances do. It is very important to know that 1) these instances exist, and 2) what features they have or where they lie in the instance space. This is a crucial contribution, because these instances pose a new challenge and this drives forward the knowledge we have about the Hamiltonian completion problem, which will hopefully ultimately result in better algorithms.
	}
	%We first evolve in instance hardness and then fill in the landscape for a more complete look at the instance space. This enables us to generate more diverse and challenging test instances than standard graph generators. \textcolor{blue}{The full set of instances are made publicly available in order to allow other researchers to benchmark their algorithms and to help develop better solvers.}
	
	We show that projecting the complex instance space down to two dimensions retains key characteristics, \textcolor{cyan}{which allows us to get direct insights into algorithm performance.}
	%that the standard generators have a limited footprint. 
	We display the power of visualization, offering several intuitions on the topics of evolutionary algorithms, algorithm performance and instance features. \textcolor{blue}{We gain useful insights into the differences between \textit{MSLS} and \textit{Concorde}, for example that certain features or properties can predict their performance.}
	To the best of our knowledge, we are the first to examine the instance space of the Hamiltonian completion problem \textcolor{blue}{and we show that existing problem instances in the literature do not cover large areas of the instance space, whereas our evolved instances do.}
	
	The remainder of this paper is structured as follows. In Section \ref{sec:HCP} the Hamiltonian completion problem and its conversion to TSP is briefly illustrated. In Section \ref{sec:Evolving} we explain the process of evolving instances. Section \ref{sec:visual} examines the visualization of the instance space, which is then further discussed in Section \ref{sec:results}. Related work and its relation to our paper is discussed in Section \ref{sec:related}. Finally, a conclusion and possible avenues for further research are explored in Section \ref{sec:conclusion}.
	
	\section{Hamiltonian completion problem}
	\label{sec:HCP}
	In order to understand the problem at hand, we must first introduce some definitions. An undirected graph is defined as a set of nodes $V$ and a set of edges $E$, where all edges can be traversed in both directions. A Hamiltonian graph is a graph that has at least one Hamiltonian cycle, which is a cycle that contains every node of the graph exactly once. 
	The Hamiltonian completion problem (HCP) on undirected graphs is as follows: given an undirected graph $G = (V,E)$, find a set of edges $E'$ such that $G'=(V,E\cup E')$ is a Hamiltonian graph and the size of $E'$ is as small as possible.
	
\begin{figure*}[h]
	\centering
	\begin{subfigure}{0.3\linewidth}
		\centering
%		\captionsetup{width=.8\linewidth}
%		\begin{tikzpicture}[auto ,node distance =1.5 cm ,on grid,
%		line width= 2.0pt ,scale=0.9, every node/.style={scale=0.9},
%		state/.style ={ circle ,top color =white , bottom color = processblue!20 ,
%			draw,processblue , text=blue , minimum width =1 cm}]
%		\node[state] (C) {$3$};
%		\node[state] (A) [above left=of C] {$1$};
%		\node[state] (B) [above right =of C] {$2$};
%		\node[state] (D) [below =of C] {$4$};
%		\node[state] (E) [below left =of D] {$5$};
%		\path (C) edge  node[below =0.15 cm] {} (A);
%		\path (A) edge  node[above] {} (B);
%		\path (B) edge  node[below =0.15 cm] {} (C);
%		\path (D) edge node[] {} (C);
%		\path (E) edge node[] {} (D);
%		\end{tikzpicture}
	
	%.65\textwidth for final
	\includegraphics[width=.65\textwidth,center]{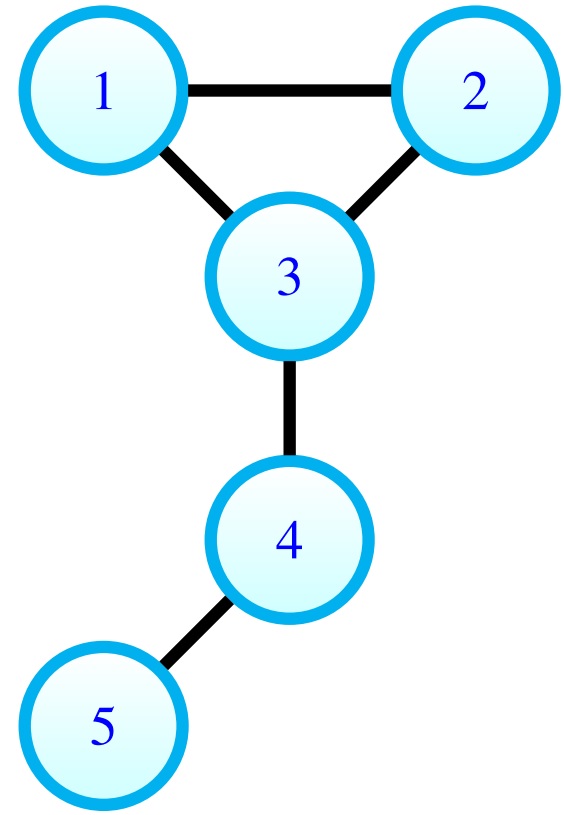}
	\caption{Graph $G$}
	\end{subfigure}
	\begin{subfigure}{0.3\linewidth}
		\centering\captionsetup{width=.8\linewidth}
%		\begin{tikzpicture}[auto ,node distance =1.5 cm ,on grid,
%		line width= 2.0pt ,scale=0.9, every node/.style={scale=0.9},
%		state/.style ={ circle ,top color =white , bottom color = processblue!20 ,
%			draw,processblue , text=blue , minimum width =1 cm}]
%		\node[state] (C) {$3$};
%		\node[state] (A) [above left=of C] {$1$};
%		\node[state] (B) [above right =of C] {$2$};
%		\node[state] (D) [below =of C] {$4$};
%		\node[state] (E) [below left =of D] {$5$};
%		\path (C) edge  node[below =0.15 cm] {} (A);
%		\path (A) edge  node[above] {} (B);
%		\path (B) edge  node[below =0.15 cm] {} (C);
%		\path (D) edge node[] {} (C);
%		\path (E) edge node[] {} (D);
%		\path (E) edge [color=red] (A);
%		\end{tikzpicture}		

		%.65\textwidth for final
		\includegraphics[width=.65\textwidth,center]{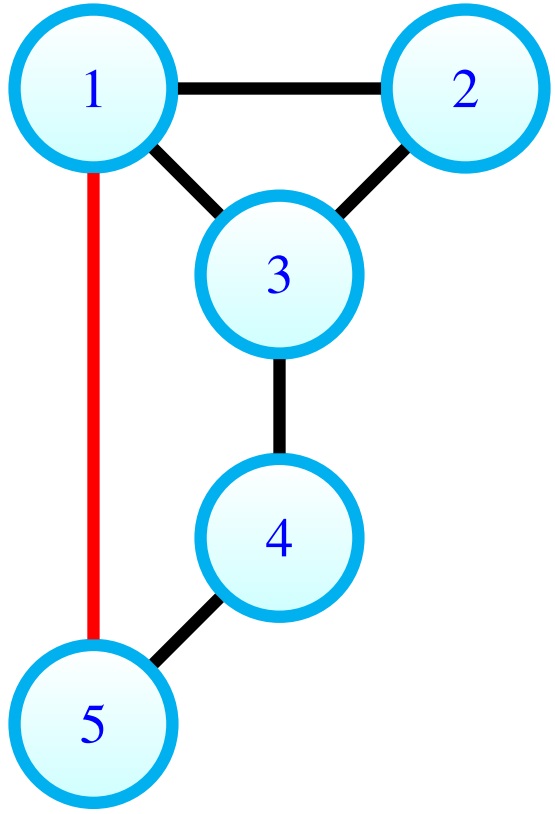}
		\caption{Graph $G'$}
	\end{subfigure}
	\begin{subfigure}{0.3\linewidth}
		\centering\captionsetup{width=\linewidth}
%		\begin{tikzpicture}[auto ,node distance =1.5 cm ,on grid,
%		line width= 2.0pt ,scale=0.9, every node/.style={scale=0.9},
%		state/.style ={ circle ,top color =white , bottom color = processblue!20 ,
%			draw,processblue , text=blue , minimum width =1 cm}]
%		\node[state] (C) {$3$};
%		\node[state] (A) [above left=of C] {$1$};
%		\node[state] (B) [above right =of C] {$2$};
%		\node[state] (D) [below =of C] {$4$};
%		\node[state] (E) [below left =of D] {$5$};
%		\path (C) edge   (A);
%		\path (A) edge [color = green!60!gray]  (B);
%		\path (B) edge [color = green!60!gray] (C);
%		\path (D) edge [color = green!60!gray] (C);
%		\path (E) edge [color = green!60!gray] (D);
%		\path (E) edge [color = green!60!gray] (A);
%		\end{tikzpicture}

		%.65\textwidth for final
		\includegraphics[width=.65\textwidth,center]{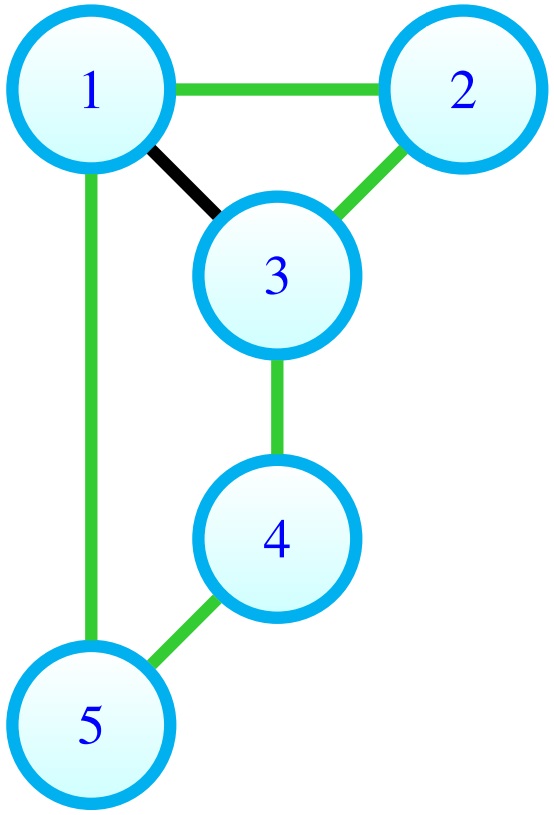}
		\caption{The Hamiltonian cycle in $G'$}
	\end{subfigure}
	\caption{Example of the Hamiltonian Completion Problem}
	\label{fig:hcpexample}
	\end{figure*}

	As an example, consider the following illustrations. Figure \ref{fig:hcpexample} (a) shows a graph instance $G$ for which HCP will be solved. There is no Hamiltonian cycle in $G$, so we add an edge between node $1$ and $5$, and call the resulting graph $G'$ (Figure \ref{fig:hcpexample} (b)). We find that there is now a Hamiltonian cycle, namely the cycle $(1-2-3-4-5-1)$ (Figure \ref{fig:hcpexample} (c)). We have added 1 edge, which results in a Hamiltonian graph and represents an optimal solution to the problem.
	
	%2 column figure
	
	\subsection{Conversion to traveling salesman problem}
	The traveling salesman problem is a minimization problem that seeks the shortest route that passes through every node of a graph exactly once and returns to the starting node. In the symmetric variant, the distance between two nodes is the same in each direction, forming an undirected graph.
	\cite{reductionTSP} state that HCP on undirected graphs can also be solved by first converting the graph to an instance of the symmetric traveling salesman problem. Let $G = (V,E)$ be an undirected, unweighted graph. Let $G' = (V',E')$ be an undirected, weighted and complete graph such that $V' = V$ and $E'$ consists of all possible edges in the graph ($|V| \times |V-1| \,/ \,2 $ edges). The weight of an edge $e \in E'$ is 0 if and only if there is an edge in $G$ that links the same pair of nodes. Otherwise the weight is set to 1. 
	HCP on $G$ can then be solved by determining the minimum weight Hamiltonian cycle in $G'$ and adding to $G$ those edges of the cycle that have a weight of 1.
	Determining this cycle in $G'$ is an instance of the symmetric traveling salesman problem. 
	With this knowledge, we can use existing TSP solvers such as \textit{Concorde} (an exact solver) \citep{Concorde} or \textit{LKH} (a heuristic solver) \citep{lkh} to solve HCP. In the following sections we compare the performance of the multi-start local search algorithm (\textit{MSLS}) \citep{jooken2019multistart} to the performance of \textit{Concorde} on the converted instance.
	
	\textcolor{blue}{
	While there is a conversion possible to TSP (so the step from TSP-diversity research might appear small), only a very small fraction of TSP instances represent HCP instances. To the best of our knowledge, most of the similar work done around TSP is focused on the 2D Euclidean TSP (with weights mapped onto the distance between nodes), but when converting HCP to TSP, only weights of 0 and 1 are possible. Therefore none of the TSP instances considered by e.g. \cite{Evolvinginstances}, \cite{milesUnderstandingTSP}, \cite{EvolvedInstancesTSP} and \cite{maxperformance} are also HCP instances. As such, we believe that specifically studying HCP instances is interesting and could even add to TSP diversity.
	}
	
	\section{Evolving instances}
	\label{sec:Evolving}
	Our goal in this section is to generate graph instances that are easy for one algorithm and hard for the other. We do this with the intention of getting a better look at their respective strengths and weaknesses. To generate these graphs, we used an evolutionary algorithm that tries to maximize the difference in runtime of the two algorithms, namely \textit{Concorde} and \textit{MSLS}.
	\textcolor{cyan}{ 
	This is a similar approach to e.g. \cite{maxperformance,maxperformance2}.}
	All experiments were performed on an Intel i7-8750H CPU with a clock rate of 2.20GHz \textcolor{blue}{and ran for a total of 160 CPU hours.} We also used the same parameters for \textit{MSLS} as suggested in \cite{jooken2019multistart}. We used Python and the DEAP library \citep{DEAP} to implement the evolutionary algorithm. \textcolor{blue}{All the resulting graphs} \textcolor{cyan}{and the code used to generate these are available on GitHub\footnote{https://github.com/thibaultLe/EvolvingTestInstancesHCP}}.
	An important thing to mention is that we are comparing a heuristic algorithm (\textit{MSLS}) with an exact algorithm (\textit{Concorde}). 
	This means that for some graphs (and specifically for graphs with more than several thousands of nodes), \textit{MSLS} might not return the optimal solution.
	\textcolor{blue}{At the same time, \textit{Concorde} might not return a solution at all within the desired time constraints (e.g. 1000 seconds). This is where using \textit{MSLS} is most useful as it can quickly find an acceptable solution in those cases. Out of the 172 problem instances considered in \cite{jooken2019multistart}, \textit{MSLS} was able to find the optimal solution for 140 instances. For our smaller yet more diverse instances, we assume (and later verify) that \textit{MSLS} is always able to find the optimal solution. Therefore we will mostly focus on the difference in runtime and not on the optimality of the solutions.}
	
	%However, for the vast majority of tested instances in \cite{jooken2019multistart}, if \textit{MSLS} did not return the optimal solution, \textit{Concorde} solved it faster. 
	%Therefore we will mostly focus on the difference in runtime and not on the optimality of the solutions.
	
	We limited the graph instances to a fixed number of nodes ($|V| = 50$) in order to make sure the runtimes were computationally manageable. 
	To illustrate, the median runtime of each algorithm on such a graph instance was around one second and we expected to need more than 10,000 instances to get a decent look at the instance space.
	This also made it possible to easily compare instances, as the only variables are the number of edges and how these are distributed in the graph. 
	\textcolor{blue}{While these instances might appear rather small, they are quite similar in size compared to other work. For example, \cite{Evolvinginstances} and \cite{EvolvedInstancesTSP} looked at TSP instances with 100 nodes, and \cite{featurediversity} examined ones with 25, 50 and 100 nodes. This is due to the inherent trade-off between instance size and number of instances that can be examined. For example, when maximizing the difference in runtime, we necessarily need to evaluate the performance of both algorithms on each instance. Examining larger instances would therefore result in being limited in how far we can evolve and analyze the instance space, as there are computational constraints.
	To the best of our knowledge, a way to efficiently evolve large problem instances has not yet been proposed.  
	}

\begin{figure*}[h]
		\centering
		\begin{subfigure}{0.21\linewidth}
%			\begin{tikzpicture}[auto ,node distance =1.5 cm ,on grid,
%			line width= 1.5pt ,scale=0.9, every node/.style={scale=0.9},
%			state/.style ={ circle ,top color =white , bottom color = processblue!20 ,
%				draw,processblue , text=blue , minimum width =1 cm}]
%			\node[state] (C) {$3$};
%			\node[state] (B) [above left =of C] {$2$};
%			\node[state] (A) [below left=of B] {$1$};
%			\node[state] (D) [below =of C] {$4$};
%			\node[state] (E) [below =of A] {$5$};
%			\path (C) edge  node[below =0.15 cm] {} (A);
%			\path (B) edge  node[below =0.15 cm] {} (C);
%			\path (D) edge (C);
%			\path (A) edge (D);
%			\path (E) edge (A);
%			\end{tikzpicture}

			%.7\textwidth, right for final
			\includegraphics[width=.9\textwidth,right]{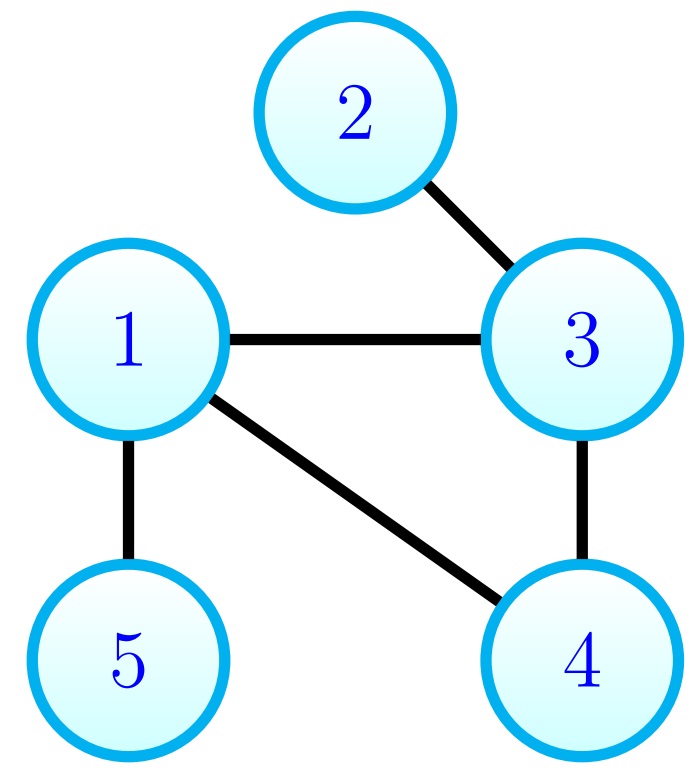}
		\end{subfigure}
		\begin{subfigure}{0.3\linewidth}
%			\begin{equation*}
%			\rightarrow
%			G =
%			\renewcommand{\arraystretch}{1.5}
%			\begin{pmatrix}
%			\multicolumn{1}{c|}0 & 0 & 1 & 1 & 1\\\cline{2-2}
%			0 &\multicolumn{1}{c|} 0 & 1 & 0 & 0\\\cline{3-3}
%			1 & 1 & \multicolumn{1}{c|}0 & 1 & 0\\\cline{4-4}
%			1 & 0 & 1 &\multicolumn{1}{c|} 0 & 0\\\cline{5-5}
%			1 & 0 & 0 & 0 & 0\\
%			\end{pmatrix}
%			\rightarrow
%			\end{equation*}		

			%\textwidth for final
			\includegraphics[width=.9\textwidth,center]{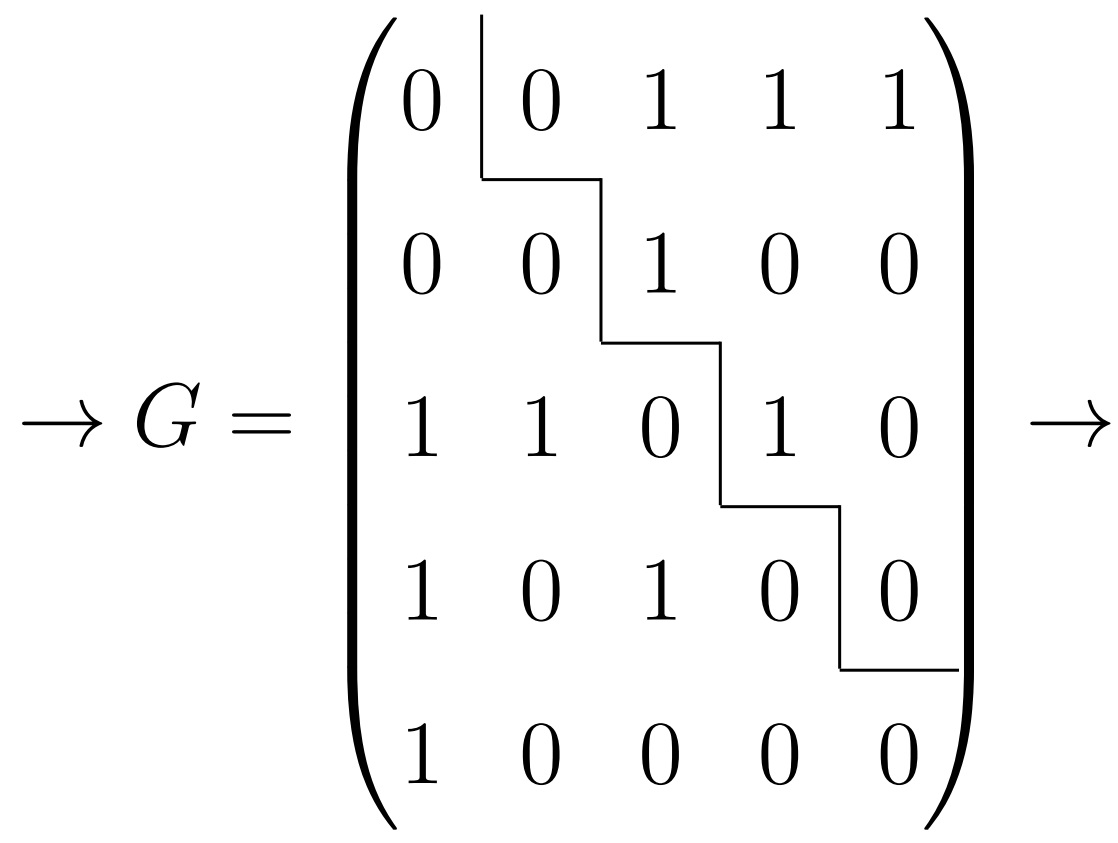}
		\end{subfigure}
		\begin{subfigure}{0.33\linewidth}
%			\begin{equation*}
%			\renewcommand{\arraystretch}{1.5}
%			\begin{pmatrix}
%			 0 & 1 & 1 & 1 & \cdots
%			\end{pmatrix}
%			\end{equation*}	

			%.65\textwidth, left for final
			\includegraphics[width=.9\textwidth,left]{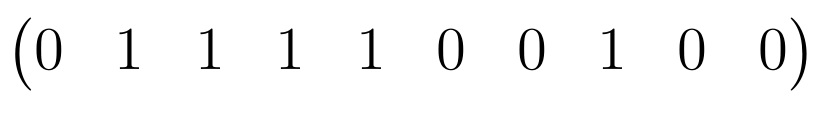}
		\end{subfigure}
		\caption{Graph encoding for use in the evolutionary algorithm}
		\label{encoding}
	\end{figure*}

	Encoding graph instances for use in the evolutionary algorithm was done by a binary vector of length $|V| \times |V-1| \,/ \,2 $, which is the total number of possible edges in a graph. The vector indicates for each possible edge whether it is present in the graph or not. The elements of the vector are therefore the upper triangular part of the adjacency matrix of the graph after unrolling, as can be seen in Figure \ref{encoding}. As HCP is solved on undirected and unweighted graphs, the binary vector holds all the information necessary to encode an instance. 

	%2 column figure

	\subsection{Evolutionary process}
\begin{figure} [h]
		\centering
%		\begin{tikzpicture}[->,node distance =1 cm]
%		\node (A) [draw, rounded rectangle] {Starting population of 20 graphs};
%		\node (C) [draw, rounded rectangle,below =of A] {Create 30 offspring};
%		\node (D) [draw, rounded rectangle,below =of C] {Evaluate the offspring};
%		\node (E) [draw, rounded rectangle,right =of D] {Select 20 graphs};
%		\node (F) [draw, rounded rectangle, above =of E,text width=3cm,align=center] {Max number of generations reached?};
%		\node (G) [draw, rounded rectangle, right =of F] {Stop};
%		
%		\path (A) edge [-{Latex[length=3mm, width=2mm]}] (C);
%		\path (C) edge [-{Latex[length=3mm, width=2mm]}] (D);
%		\path (D) edge [-{Latex[length=3mm, width=2mm]}] (E);
%		\path (E) edge [-{Latex[length=3mm, width=2mm]}] (F);
%		\path (F) edge [-{Latex[length=3mm, width=2mm]}] node[above = 0.04cm] {No} (C);
%		\path (F) edge [-{Latex[length=3mm, width=2mm]}] node[above = 0.01cm] {Yes} (G);
%		\end{tikzpicture}

		%.65\textwidth for final
		\includegraphics[width=\linewidth,center]{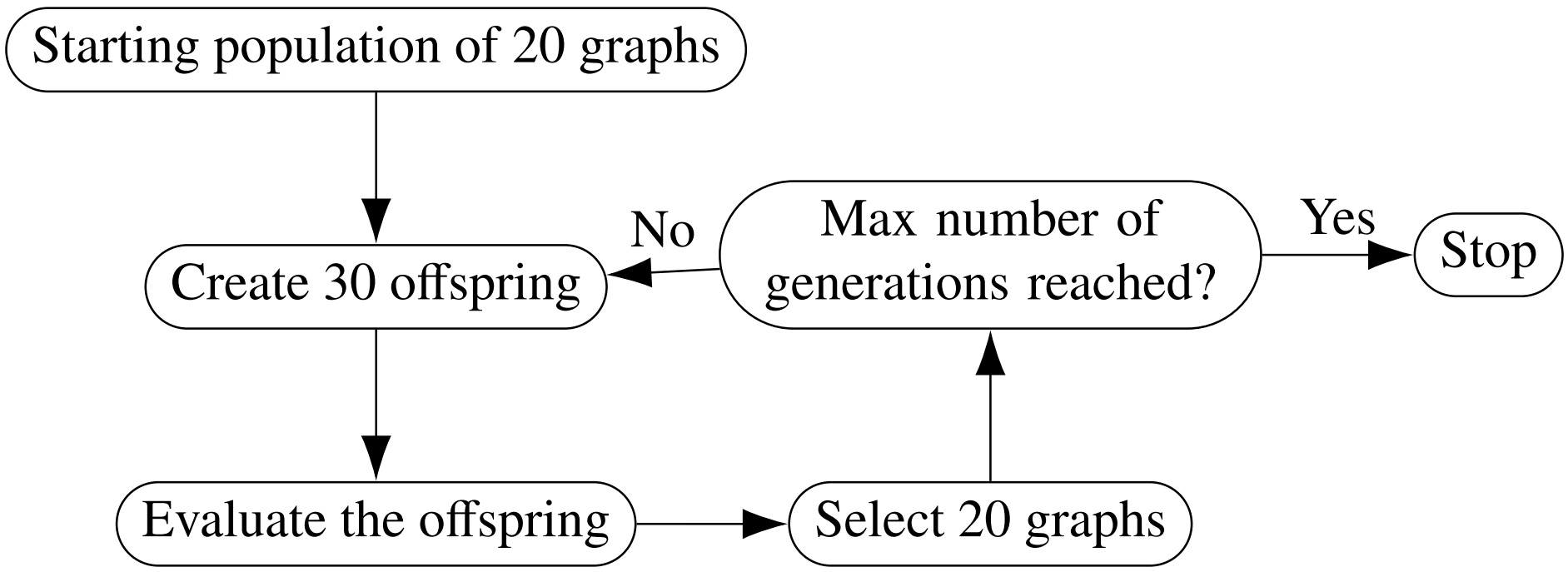}
		\caption{The process of evolving graph instances}
		\label{evoprocess}
	\end{figure}

	We started each run of the evolutionary algorithm with a randomly generated starting population of 20 graphs. These are Erdős-Rényi graphs \citep{erdos}, 
	%where each pair of nodes is connected by an edge with probability $p$.
	which are constructed by randomly connecting nodes. Each edge is included in the graph with probability $p$.
	This forms the first step in the evolutionary process, depicted in Figure \ref{evoprocess}. 
	%Every run had a different probability value, with edge counts varying from 5 to 1225, the latter being the maximum number of edges for a graph of 50 nodes. 
	The probabilities were chosen such that the edge counts vary between 5 and 1225 (the maximum number of edges for a graph of 50 nodes).
	This enabled us to gain insight into the performance of the two algorithms for the full range of graph densities. 
	%Then we enter the evolutionary loop, which generates 30 new graphs from the current population. 	
	Next, the evolutionary loop is entered, in which 30 new graphs are generated based on the current population.
	This is done in three ways: each of the 30 times, either a graph from the current population is cloned, either a graph is mutated or two graphs are merged using two-point crossover. Which one of the three happens is randomly chosen, each having the same probability. 
	When a graph is mutated, up to 3\% of edges are added, removed or replaced by other edges. This is achieved by flipping values of the binary vector encoding of a graph with a probability of 3\%. When crossover is selected, two children are generated using two-point crossover on the vector encoding of the parents. The crossover points are picked randomly. %which is equivalent to performing two single-point crossovers with different crossover points. 
	One of these two children is selected at random and added to the offspring.
	
	These 30 graphs are then evaluated by using both \textit{Concorde} and \textit{MSLS} to solve HCP on each graph. The difference in runtime of the two algorithms is recorded and used as a fitness function. Process-specific time was used in order to limit the effects of task preemption by the operating system. 
	The fitness function is then defined as follows: runtime(\textit{MSLS})~-~runtime(\textit{Concorde}). When maximizing this function, we evolve towards graphs where \textit{Concorde} is faster. When minimizing, we evolve towards graphs where \textit{MSLS} is faster. 
	
%	\begin{center}
%		\begin{tabular}{ |c|c|c|c|c|c|c|c| } 
%			\hline
%			0 & 1 & 1 & 1 & 1 & 0 & 0 & 1\\ 
%			\hline
%		\end{tabular}
%	\end{center}

\begin{figure}[h]
		%\centering
		%Both single column images
		\begin{subfigure}{\linewidth}
			\centering
			\includegraphics[width=.8\textwidth,center]{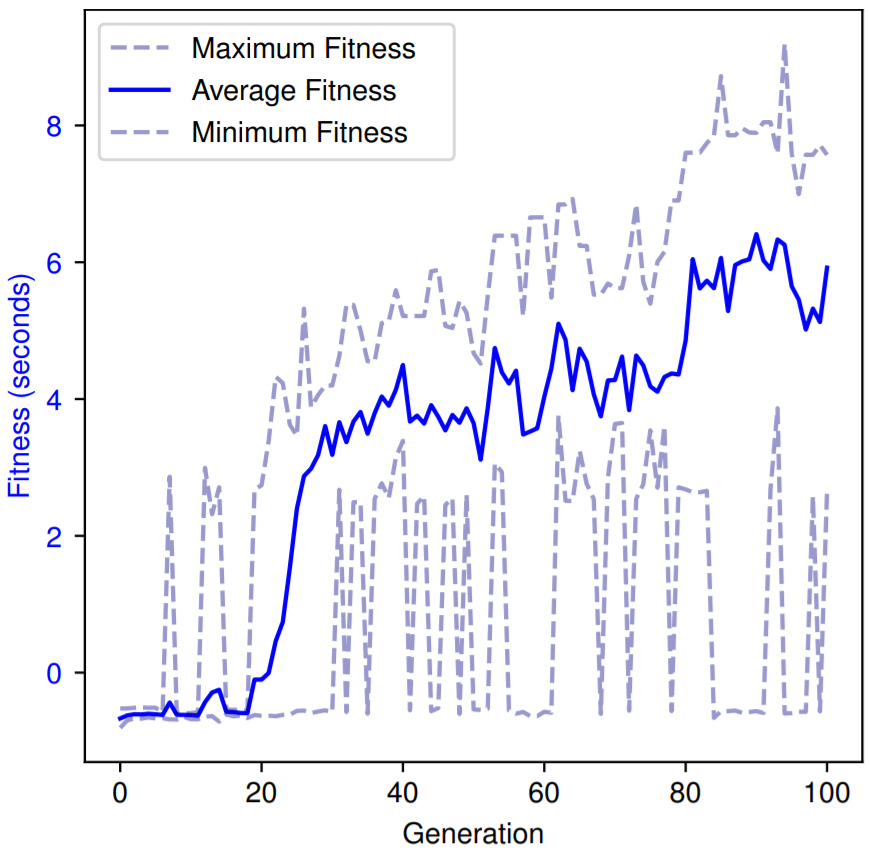}
			\caption{The fitness of the population}
		\end{subfigure}
		\begin{subfigure}{\linewidth}
			\centering
			\includegraphics[width=.8\textwidth,center]{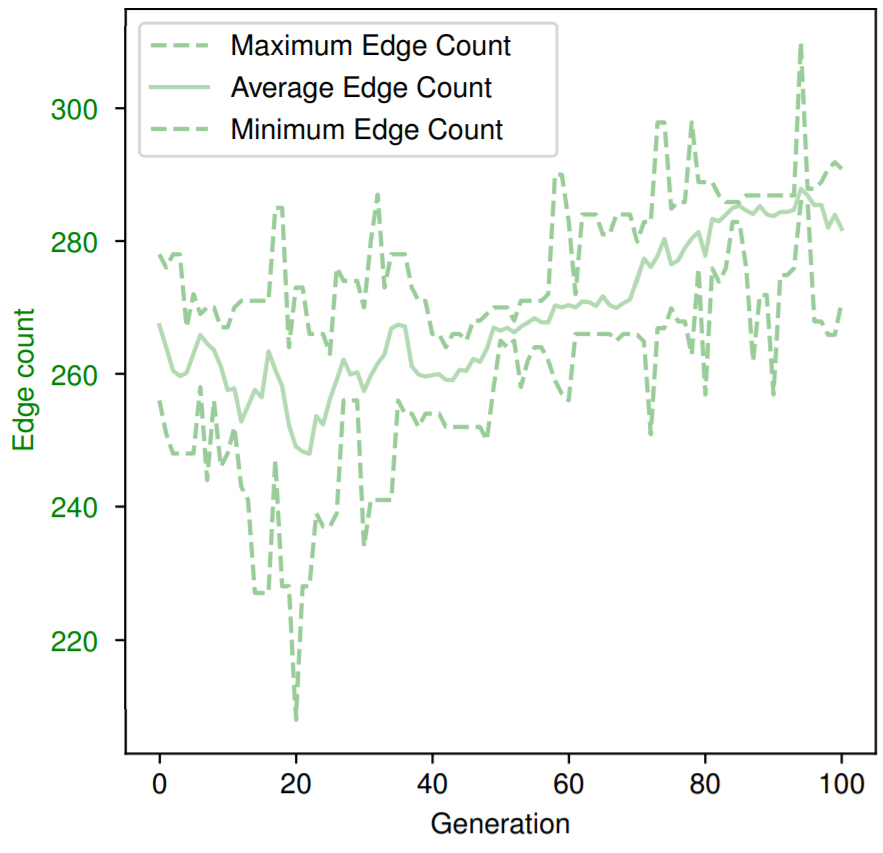}
			\caption{The number of edges of the population}
		\end{subfigure}
		\caption{Statistics of a sample run of the evolutionary algorithm that maximizes fitness}
		\label{fig:fitnessevo}
	\end{figure}

	After having evaluated the offspring, 20 graphs are selected through 2-tournament selection \citep{tournament} to become the new population and the evolutionary loop continues. Thus the size of the population is kept constant at 20 individuals. When reaching 100 generations, the  loop is stopped. A 'hall of fame' is also recorded, which kept track of the 300 best individuals that ever existed, together with their fitness values. 
	As an example, the fitness values of the population of a run of the evolutionary algorithm were plotted in Figure \ref{fig:fitnessevo}, where the algorithm aims to maximize the fitness function. This is one of 15 different runs, each with a different starting population and either maximizing or minimizing the fitness function. We can see that the minimum, average and maximum fitness values of the population of this run evolves throughout the generations. The average number of edges of the population is also shown as a reference. 
	
	If the average fitness values did not stagnate, but instead exhibited a rising trend during the final generations, we continued the generational loop for 100 extra generations to a total of 200 generations.
	This is present in the previously mentioned figure, thus we continued the loop for this example. This led to even higher fitness values, namely a maximum of 165 seconds instead of the 8 seconds reached in the first 100 generations. Increasing the generational limit again did not result in better fitness values.
	
	\textcolor{cyan}{In order to test whether starting the evolutionary process from non-Erdős-Rényi graphs is useful, we did multiple long runs where we compared starting from Erdős-Rényi graphs to starting from preferential attachment graphs generated using the Barabási–Albert model \citep{barabasi}. There were slight differences, but they seemed to converge to the same local optima in terms of difference in algorithm performance and the deviation was not much larger than simply running the evolutionary algorithm twice. We refer to \ref{fig:EvoPA} and \ref{fig:EvoER} for the figures which support these statements. We therefore consider the difference to be small enough such that our large sample of graphs and long evolutionary runs are not missing crucial parts of the instance space.}
	%Sometimes we saw a rising trend in fitness during the final generations. This means that if we continued evolving, we would probably find graphs with even higher fitness values. Therefore we continued the generational loop for 100 extra generations when we saw this phenomenon. 
	%This is also present in the previously mentioned figure. We see that the fitness values have not reached a local optimum, but keep a rising trend towards the final generations. So we continued the generational loop for  100 generations (200 generations in total). We reached even higher fitness values, namely a maximum of 165 seconds instead of the 8 seconds of the first 100 generations. Increasing the generational limit again did not result in better fitness values.

	\subsection{Evolved instances}
	\label{sec:evolvedinstances}
	After running the evolutionary algorithm 15 times and saving 300 of the best individuals each time, we can analyze the resulting graphs. First, it was much easier to maximize the fitness function than it was to minimize it. When maximizing, we reached fitness values of up to 166 seconds, which means that \textit{Concorde} solved HCP 166 seconds faster than \textit{MSLS} did on the same graph. When minimizing however, we only reached values of around -4 seconds, where \textit{Concorde} solved HCP 4 seconds slower than \textit{MSLS}. Note that the runtime of \textit{Concorde} was mostly stable, ranging from around 0.5 to a few seconds to solve each graph. \textit{MSLS} on the other hand had a very wide range of runtimes, ranging from around 0.5 to 170 seconds. 
	%This could be explained by the fact that \textit{Concorde} is a state-of-the-art solver,  which is still being used in research to this day despite having been around for over twenty years. Therefore a more consistent and smaller range of runtimes is expected. 
	This means that the large difference in runtime is mostly attributed to \textit{MSLS} having trouble solving some of these graphs.
	This is discussed further in Section \ref{sec:correlationInstance}.
	
\begin{figure}
		\centering
		%1 or 1.5 column fitting image
		\includegraphics[width=\linewidth,center]{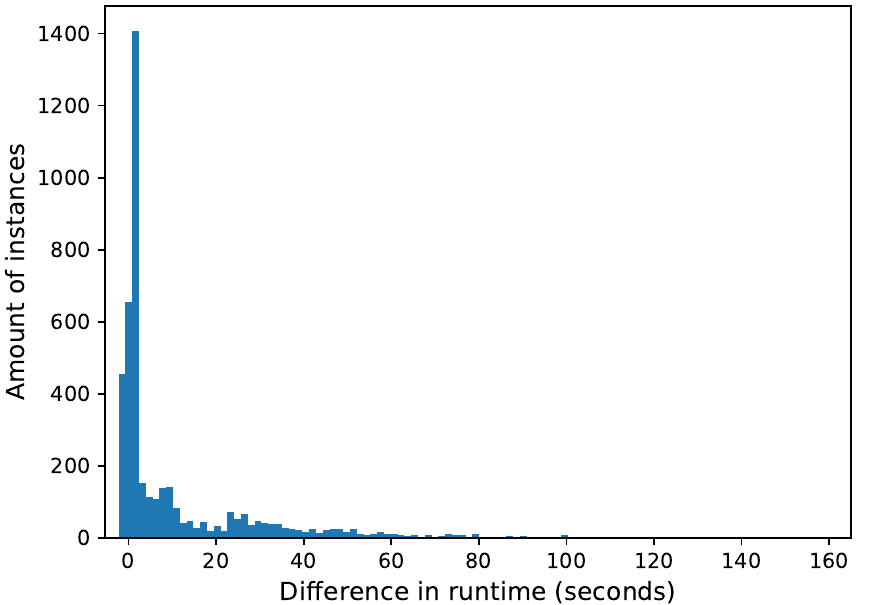}
		\caption{A histogram of the differences in runtime of the evolved instances}
		\label{fig:histogram}
	\end{figure}

	Second, we were able to evolve graphs with a much bigger difference in runtime than standard graph generators. We used the six graph generators used in \cite{jooken2019multistart} to generate 182 graphs with 50 nodes each and calculated the difference in runtime of both algorithms. These were Erdős-Rényi graphs, circle graphs, grid graphs, preferential attachment graphs, a structured tree graph and a star graph, ranging from 10 to 1220 edges. Some of these graphs were generated using the library of the Stanford
	Network Analysis Platform \citep{snap}. This resulted in a range of [-0.97,~ 2.08] seconds, which shows that our evolutionary algorithm is able to generate graphs with much larger differences in runtime, as we found a range of [-3.83, 165.55] seconds in the evolved instances. A histogram of the differences in runtime of the evolved instances is shown in Figure \ref{fig:histogram}.

	Third, the number of edges is a factor in the difference in runtime, but is definitely not the only contributing feature. The distribution of the edges in the graph is likely to be an important factor as well. We also see that small differences between two graphs can result in a large difference in runtime. For example, randomly replacing 3\% of edges can change the difference from 0.6 to 3 seconds. A more thorough analysis of why this happens is necessary, which is carried out in the next sections.

	\section{Visualizing the instance space}
	\label{sec:visual}
	In this section we will visualize the instance space of our evolved instances. As graphs have a lot of different features, it would not be very useful to just plot the number of edges and the difference in runtime in two dimensions and analyze this space. This would disregard important properties such as the edge distribution, the connectivity of the graph, the diameter and numerous other features. As we cannot directly visualize an 10-dimensional space, we need to find a way to project the high-dimensional feature space down to two dimensions. Multiple dimensionality reduction methods were considered and we chose Principal Component Analysis (PCA).
	
	\subsection{Feature space}
	Ten different features of graph instances were chosen as they had a high correlation with instance hardness in preliminary tests. These are features that relate to the edge distribution, the connectivity of the graph and many other properties. The list is as follows: the density of the graph, the clustering coefficient, the energy of the graph, the maximum degree of the nodes, the standard deviation, skewness and kurtosis of the degree distribution, the diameter, the percentage of nodes with a degree of 1 and the percentage of nodes with a degree of 2.
	
	A few definitions are useful: the density of a graph is defined as $2*|E| / (|V|*|V-1|)$, which is the number of edges divided by the total number of possible edges. The energy of a graph is defined as the sum of the absolute values of the eigenvalues of the adjacency matrix \citep{energy}. The degree of a node is the number of connections it has to other nodes. The degree distribution is then the probability distribution of these degrees. 
	%The diameter of a graph is defined as the largest number of nodes which must be traversed in order to travel from one node to another in the shortest way possible. 
	The eccentricity of a node is the largest number of nodes which must be traversed in order to travel from this node to another node in the shortest way possible. The diameter of a graph is the maximum eccentricity of any node in the graph.
	For example, the diameter of graph $G$ in Figure \ref{fig:hcpexample} (a) is 3. 
	\textcolor{blue}{Note that we also experimented with other features (e.g. ones related to the minimum spanning tree of a graph), but these were either not correlated to instance hardness, were linearly correlated with the ones we eventually selected, were too computationally expensive to calculate, or did not increase the variance of the features in the first 2 dimensions.} \textcolor{cyan}{There are also a lot of TSP features available, see e.g. footnotes 5, 6 and 7 of \cite{kerschke}. However these are almost all related to the 2-D aspect of the TSP problem, for example using the convex hull, centroid or rectangular area, which is not applicable to HCP.
	As such, we deem the used feature set to be adequate.} 
	
	\subsection{Visualization}
\begin{figure*}[h]
		\centering
		\includegraphics[width=.7\linewidth,center]{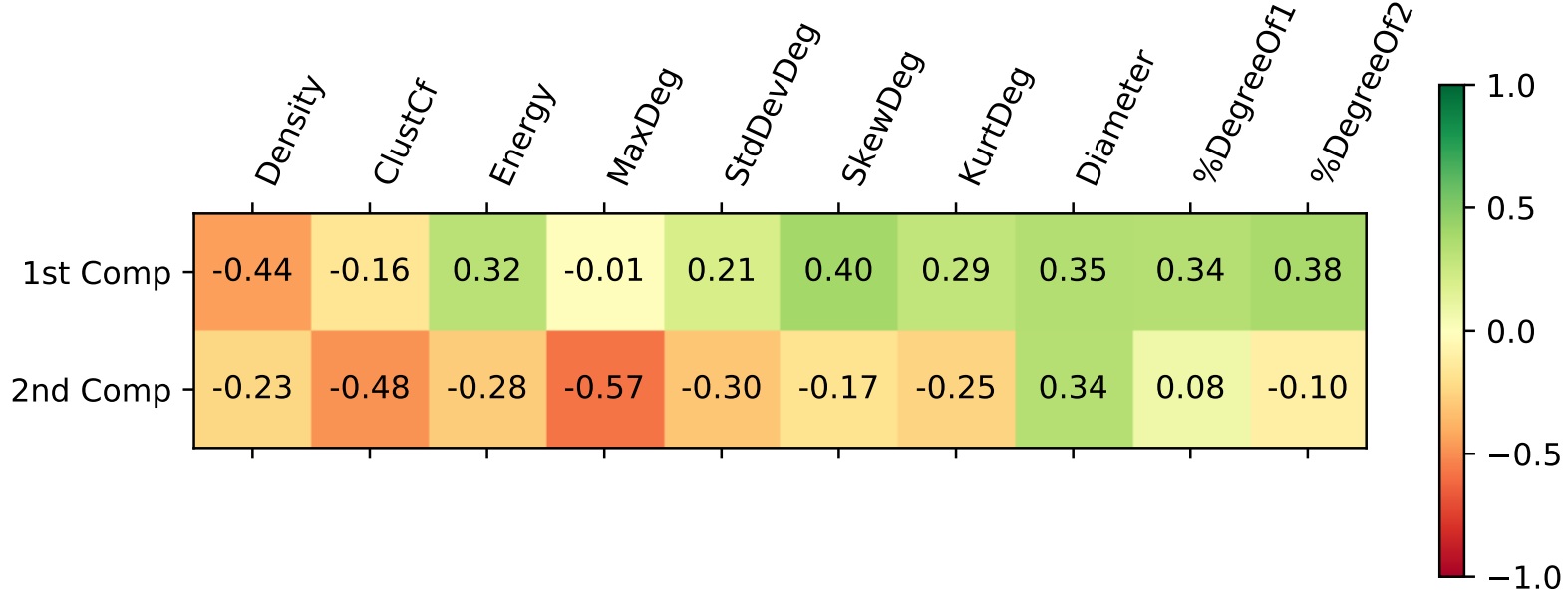}
		\caption{The coefficients of the first two principal components}
		\label{fig:pcacft}
	\end{figure*}

\begin{figure*}[h]
		\includegraphics[width=.7\textwidth,center]{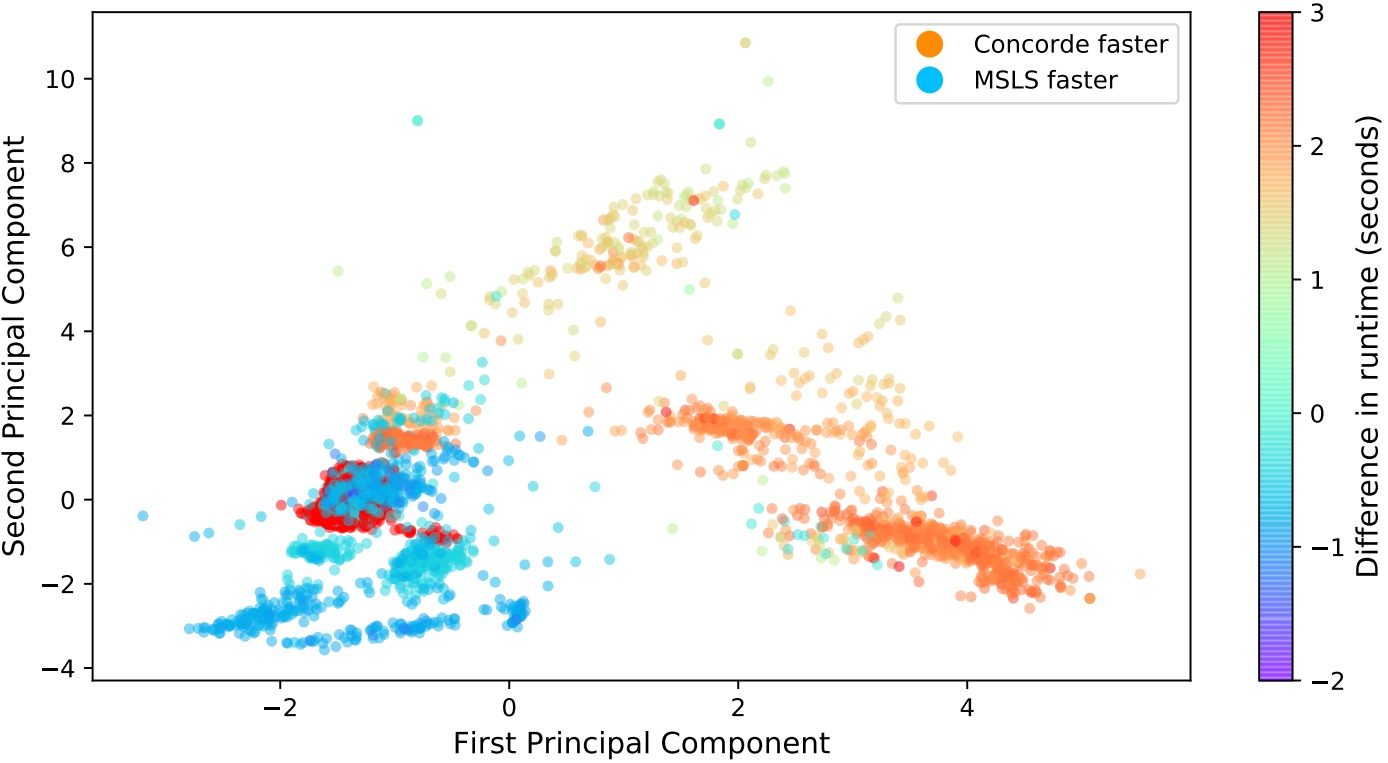}
		\caption{The landscape of the evolved instances}
		\label{fig:pca}
	\end{figure*}

	After having evolved a diverse set of 4500 instances, we are now ready to visualize the instance space using Principal Component Analysis (PCA). We also included the 182 instances generated by standard graph generators, since they have certain attributes that are not found in the evolved instances.
	As we will be visualizing the space in two dimensions, two principal components are required. The $x$-axis will be the first component and the $y$-axis will be the second component. 
	We first performed a log-transformation on all the features in order to de-emphasize outliers (negative values were handled by adding a constant value prior to the log-transformation). We then normalized the features by removing the mean and scaling to unit variance and performed the PCA on them.
	This resulted in two principal components that retained 69.0\% of the variance of the original data. The coefficients of the principal components can be seen in Figure \ref{fig:pcacft}. For example, the $x$-axis is a linear combination of the 10 features, namely $-0.44 \times $(density) $ -0.16 \times $(clustering coefficient) $ + [\ldots]$. Now we can plot our instances on these axes. For each of the 4682 instances, the 10 features are calculated, logtransformed, normalized and projected onto the $x$- and $y$-axes using the PCA coefficients. The resulting plot can be seen in Figure \ref{fig:pca}, where each dot is a different graph instance. A colormap also gives each instance a color, namely the difference in runtime of \textit{Concorde} and \textit{MSLS} solving HCP on that instance. This was the fitness value of the previous section. The differences are limited to [-2,3] in order to better show the gradient of colors, since most instances are in this range and instances with a difference higher than 3 plot to roughly the same region.

	%2 column image

	%2 column image

	\subsection{Analyzing the landscape}
	At first glance we see that there are regions in the landscape where every graph has approximately the same color. As graphs with roughly the same features are projected to roughly the same region, we see that the performance of the algorithms on similar instances can also be expected to be similar. This suggests that we can predict algorithm performance based on the location of the instance in the landscape. The second thing we can see is that there seems to be a red region in the bottom left of the landscape, with $x \in [-2,0]$ and $y \in [-1,1] $, where instances have very high fitness values but are in the same region as instances with low fitness values. This could be explained by the fact that we are projecting instances from an 10-D feature space down to two dimensions, so sometimes instances with very different features plot to the same region. It could also mean that there are some important features that can predict performance but were not included in these 10 features. This red area contains instances where \textit{MSLS} solves HCP up to 166 seconds slower than \textit{Concorde} does. 
	
	A third observation is that because we tried to either maximize or minimize the difference in runtime, we have a few regions where the fitness is either very high or very low. However, we barely have any instances in between these regions, where we might expect average fitness values or where there could be a border between the instances with positive and instances with negative fitness. In order to get a look at the entire instance space, it would be useful to fill in these empty regions before further analyzing the landscape.
	
	\subsection{Filling in the instance space}
	In the next paragraphs we will discuss two additional evolutionary algorithms in order to fill in the empty regions of the landscape, where there could be instances with certain combinations of features that we have not seen before. Note that empty regions do not always suggest that no graphs can exist there, but rather that our evolutionary algorithm did not evolve towards these areas.
	In order to fill in the sparse or empty regions, we used a second evolutionary algorithm. The idea here is that we evolve away from other graphs in the landscape. By doing so we can quickly fill up regions where there are little or no graphs, \textcolor{cyan}{similar to e.g. \cite{SMITHMILES2015102}}.
	We used the same framework as our previous evolutionary algorithm of Section \ref{sec:Evolving}, only the fitness function was changed. The fitness function is now the distance to the nearest point in the landscape and the algorithm aims to maximize this function, therefore evolving away from other points. The evaluation step in the evolutionary process in Figure \ref{evoprocess} is then as follows: calculate the 10 features of each instance, logtransform, normalize and project them on the landscape, and return the distance to their closest neighbor in the landscape as the fitness values. We ran this algorithm for 15 iterations, saving around 650 of the best instances each iteration. This resulted in 10,000 new graphs and projecting them on the landscape filled in previously sparse and empty areas. 
	
\begin{figure*}[h]
		\centering
		\includegraphics[width=.7\textwidth,center]{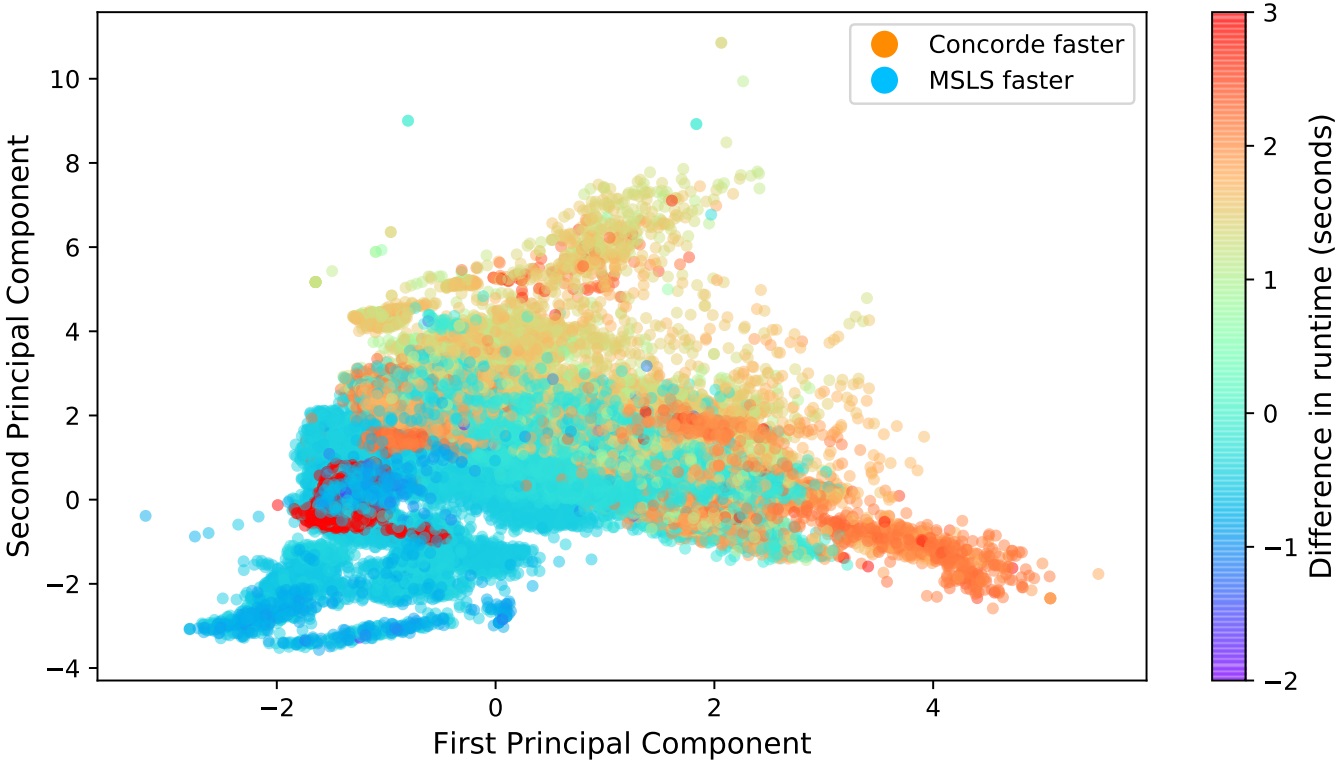}
		\caption{The filled in landscape of the evolved instances}
		\label{fig:pcafill}
	\end{figure*}

	However, there were still some interesting areas that had little to no instances.
	For example, there was a gap between instances to the left and right side of $x = 1$ (roughly between instances that are solved faster by \textit{Concorde} and ones that are solved faster by \textit{MSLS}), where we could expect a gradient of differences in runtime.
	In order to fill these specific areas more effectively, we used another fitness function, which leads to the third evolutionary algorithm. We first selected a target point in the middle of such an area, for example the point (1, 2). The fitness function is then the distance of the projected instance to the target point and the algorithm aims to minimize this function, therefore evolving towards the target point. Another 15 iterations were done with different target points, saving 650 of the best instances again, resulting in another 10,000 graphs being added. This brought the total number of graphs to 24,682, comprised of 4,500 instances of the first evolutionary algorithm, 20,000 instances of the second and third algorithm, and 182 of the standard graph generators. The resulting landscape can be seen in Figure \ref{fig:pcafill}. 
	\textcolor{cyan}{Note that we also tried to apply PCA again to this full set of instances, but this did not show new gaps in the space and only resulted in some slight stretching (see \ref{fig:pca_again}), so we kept the original PCA coefficients.}

	% 2 column fitting image

	\section{Results and discussion}
	\label{sec:results}
	In this section we provide a more thorough analysis of the full set of evolved instances. We will examine the difference in runtime and analyze correlation with certain instance features, providing a detailed look into both algorithms' strengths and weaknesses. At the end we will predict algorithm performance based on the location of an instance in the landscape.
	
	\subsection{Runtime analysis}
\begin{figure}[h]
		\centering
		\includegraphics[width=.95\linewidth,center]{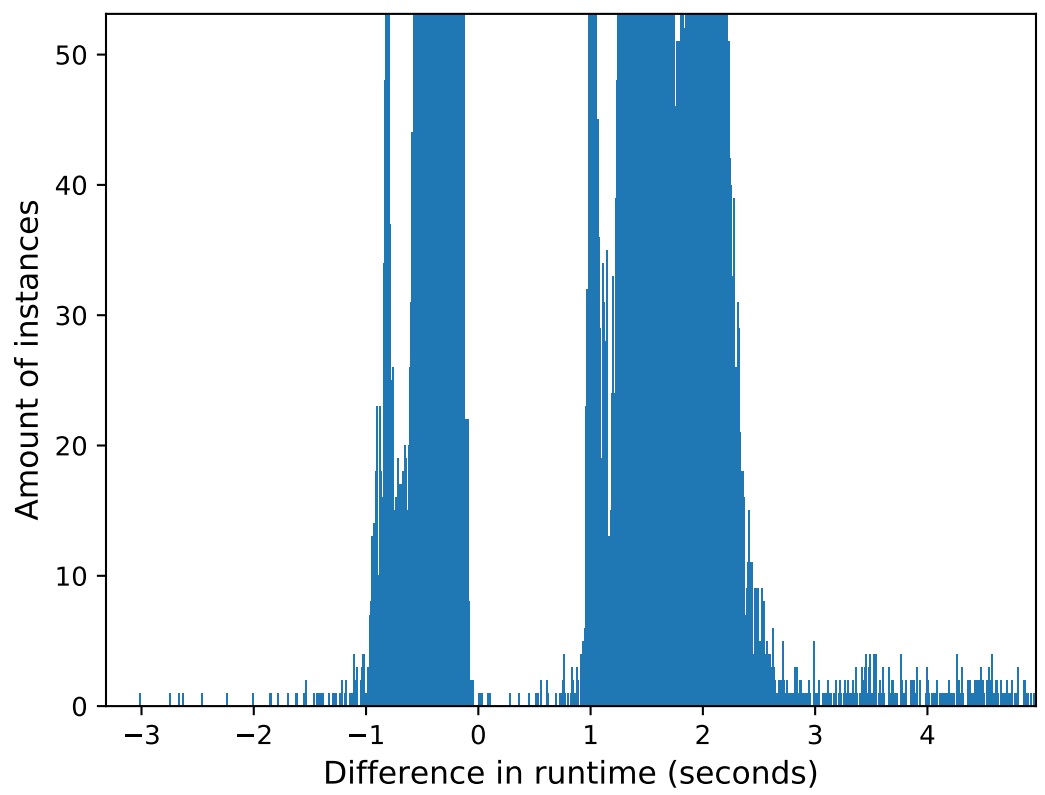}
		\caption{The zoomed-in histogram of the differences in runtime}
		\label{fig:histozoomed}
	\end{figure}

	The landscape of the evolved instances in Figure \ref{fig:pcafill} provides some interesting insights. First, there seem to be distinct areas where all instances in that area have similar differences in runtime. For example, all the instances in the bottom left region, with $x \in [-3,0]$ and $y \in [-4,-1]$, get solved roughly one second faster by \textit{MSLS}. We also see that there is an inexact border between instances that get solved faster by \textit{Concorde} and ones that get solved faster by \textit{MSLS}. Instances on this border have a roughly 50\% chance of getting solved quicker by either algorithm. However, there is not a smooth gradient of colors: the difference in runtime is either above one second or below zero seconds, with hardly any instances in between. This can be seen more clearly in Figure \ref{fig:histozoomed}, where a zoomed-in histogram of the differences in runtime is given. It implies that if \textit{Concorde} is faster, it is faster by at least one second in the large majority of instances. 
	This in an interesting result as we would expect such a histogram when maximizing the difference in runtime, but we have included over 20,000 other graphs for which we would expect a more normal distribution of differences in runtime.

	% 1 column fitting image
	
	\subsection{Footprint of standard graph generators}
\begin{figure} 
		\centering
		\includegraphics[width=\linewidth,center]{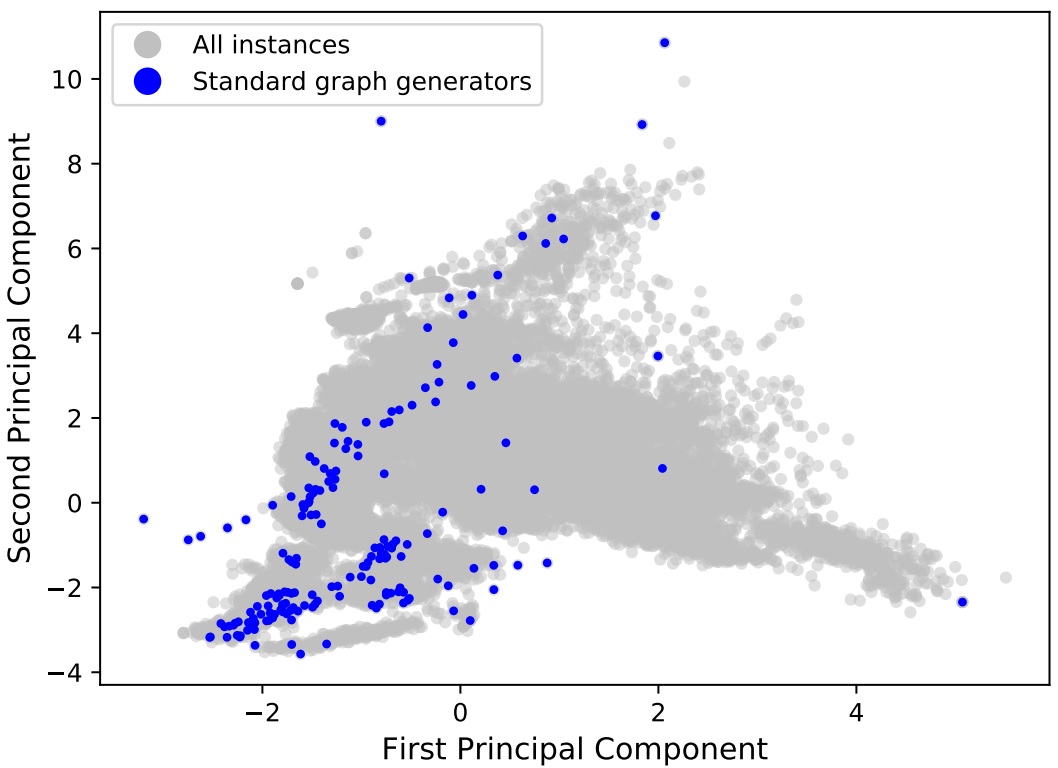}
		\caption{The footprint of standard graph generators}
		\label{fig:generators}
	\end{figure}

	An important question is whether or not standard graph generators can provide a solid set of test instances to benchmark algorithm performance. A good test set would contain diverse instances that %resemble a randomly selected sample of all the possible graphs. 
	can reveal the strengths and weaknesses of each algorithm.
	In order to observe their footprint, we can plot the graphs generated by standard graph generators as mentioned in Section \ref{sec:evolvedinstances} on our landscape. This was done in Figure \ref{fig:generators}, where the generated instances are marked in blue. We see that standard graph generators are able to reach points on the far edges of the landscape. These are specific kinds of graphs, for example circle or grid graphs, which are very hard to reach for an evolutionary algorithm using random mutation. We also see that there are large areas that are barely or not covered by the generated instances, for example the area with positive $x$-values. This can explain why they have a very limited range of differences in runtime, as they do not cover the whole landscape nor have instances with adequately diverse features.

	% 1 column fitting image

	\subsection{Correlation with instance features}
	\label{sec:correlationInstance}
\begin{figure*}
		\centering
		\begin{subfigure}{0.49\linewidth}
			\centering
			\includegraphics[width=.95\textwidth,left]{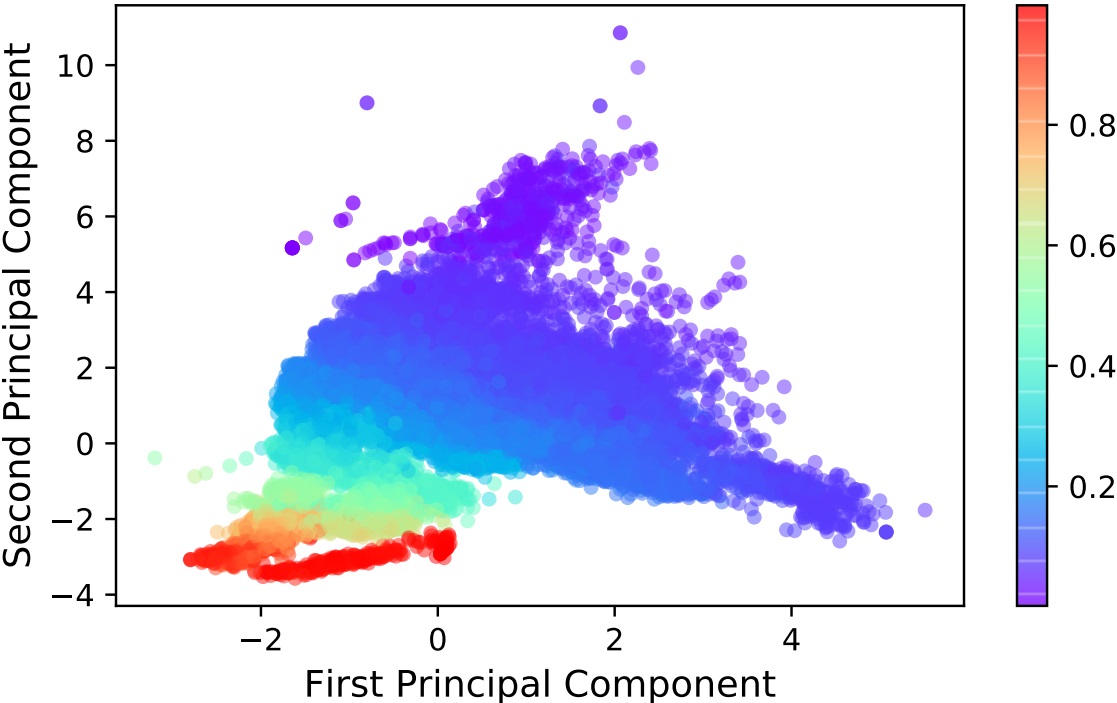}
			\caption{Density}
		\end{subfigure}
		\begin{subfigure}{0.49\linewidth}
			\centering
			\includegraphics[width=.95\textwidth,right]{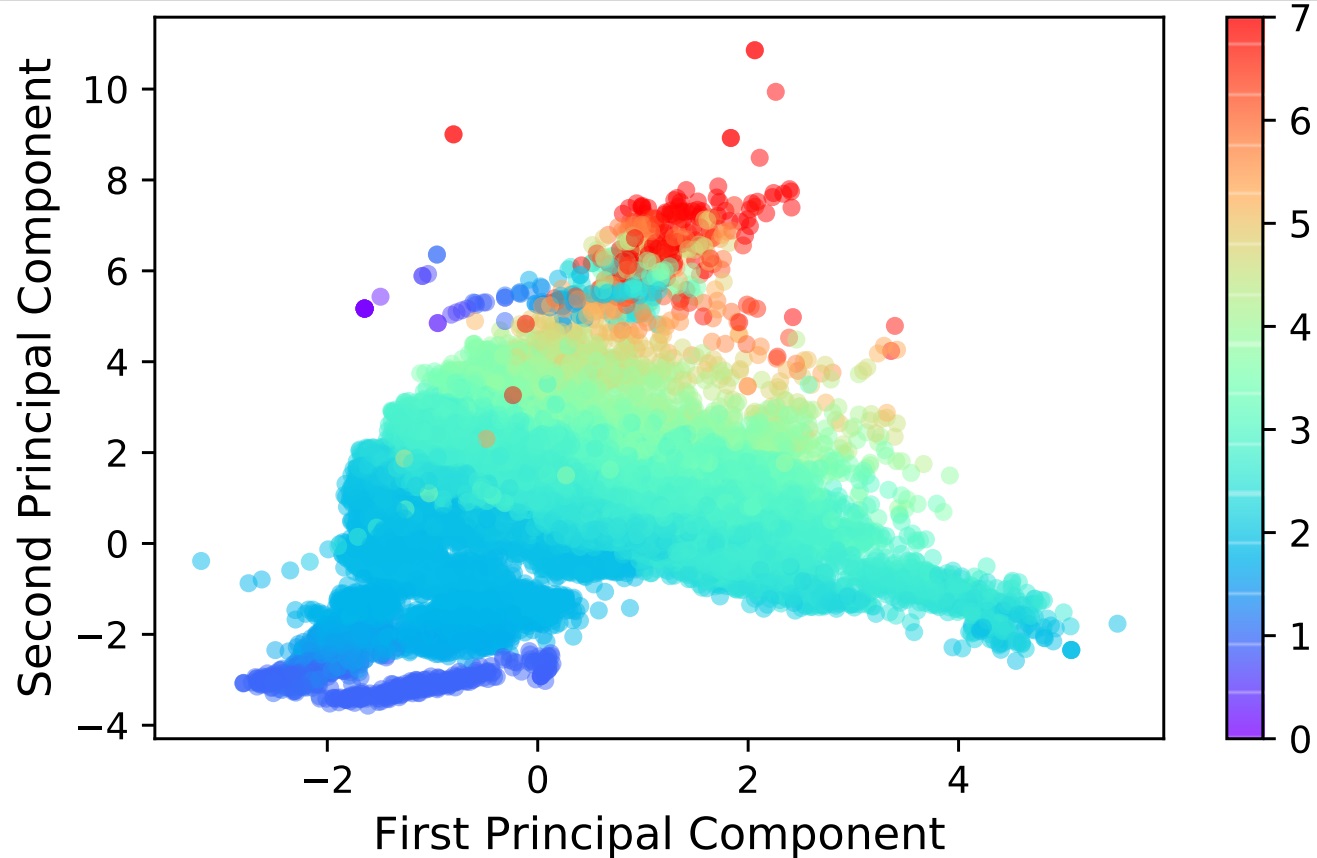}
			\caption{Diameter}
		\end{subfigure}
		\par\bigskip
		\begin{subfigure}{0.49\linewidth}
			\centering
			\includegraphics[width=.95\textwidth,left]{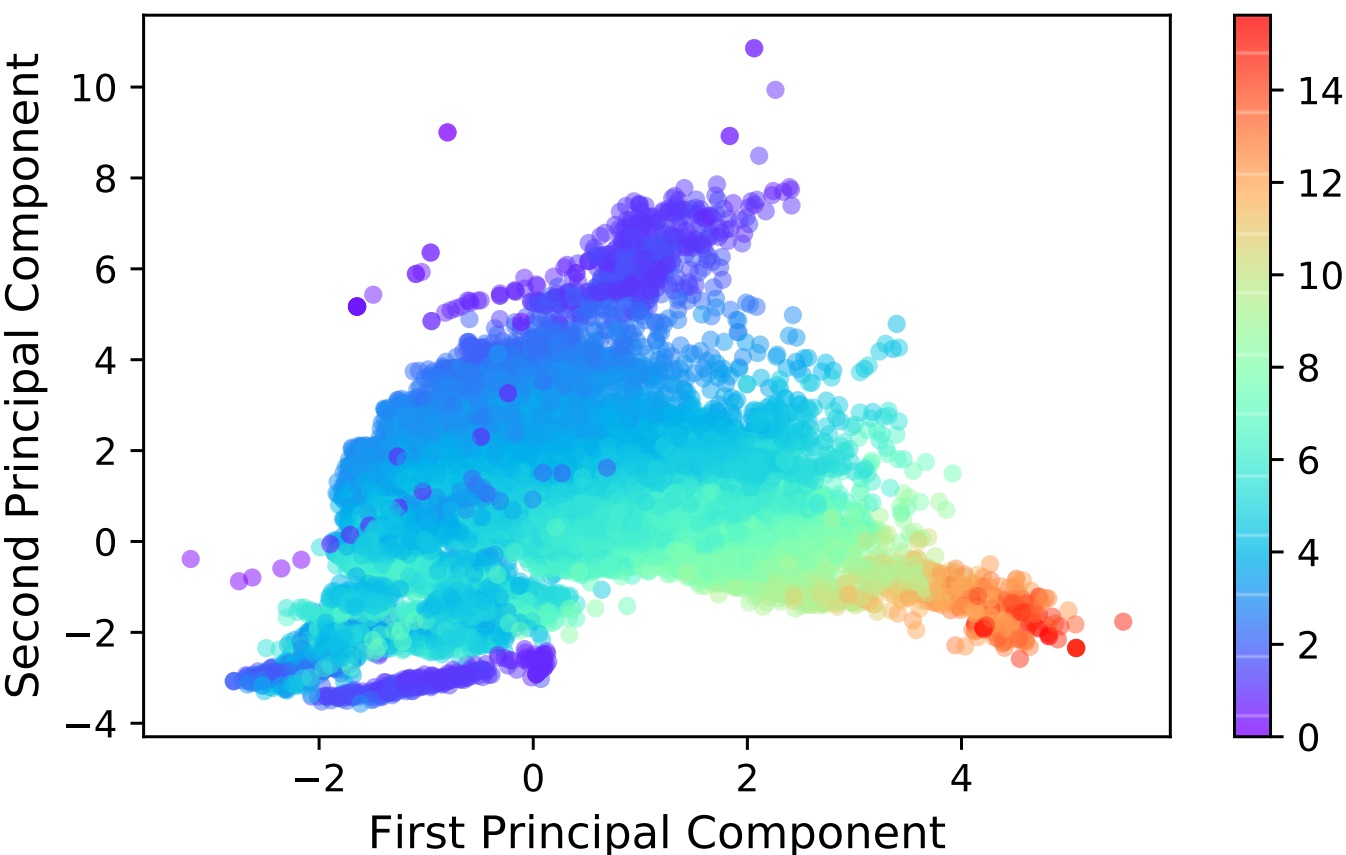}
			\caption{Standard deviation of the degree distribution}
		\end{subfigure}
		\begin{subfigure}{0.49\linewidth}
			\centering
			\includegraphics[width=.95\textwidth,right]{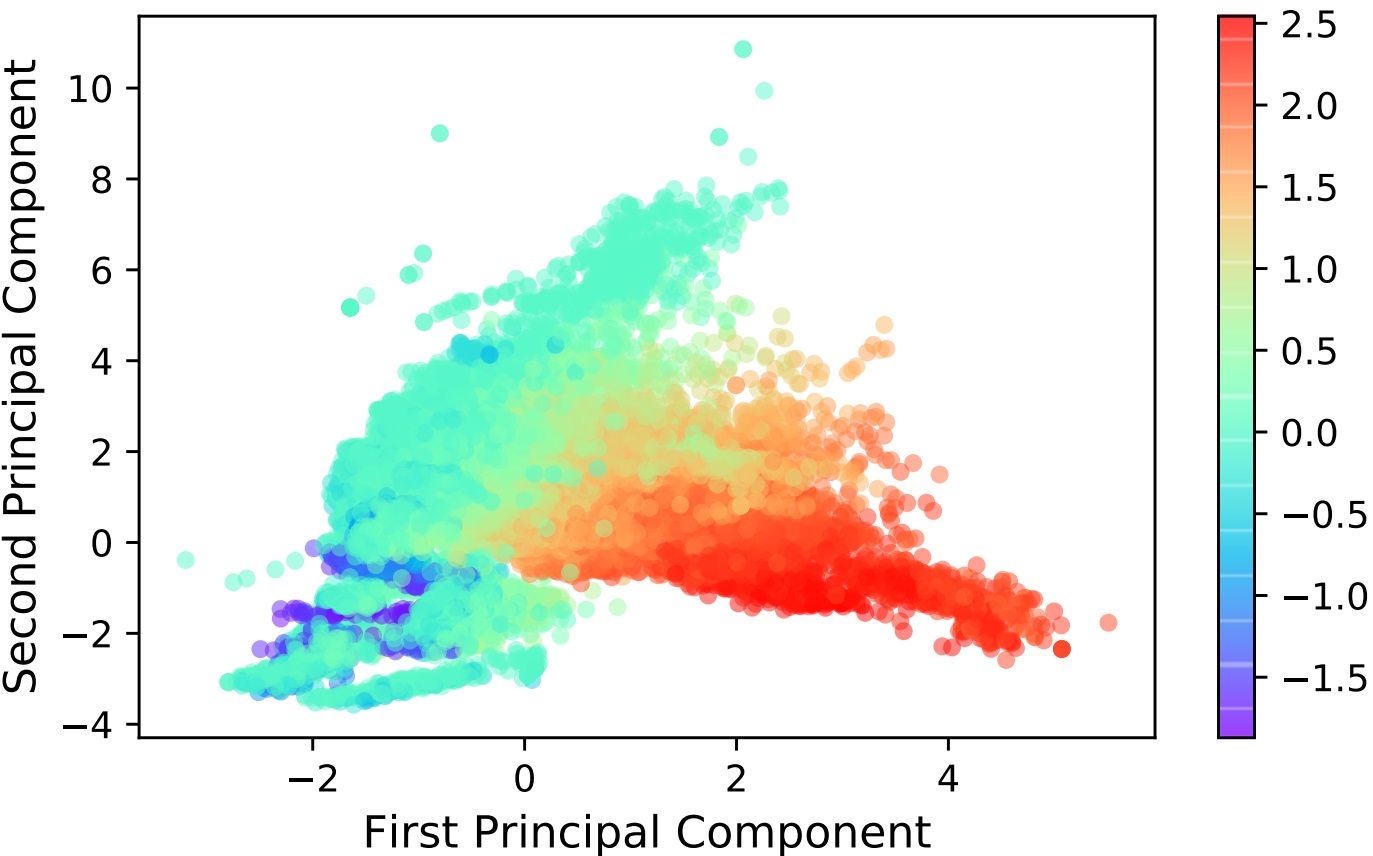}
			\caption{Skewness of the degree distribution}
		\end{subfigure}
		\caption{A selection of interesting features of instances plotted on the landscape}
		\label{fig:landfeatures}
	\end{figure*}

	Instead of using the difference in runtime to color the instances in the landscape, as was done in Figure \ref{fig:pcafill}, we can also use the features of the instances. This can give a better intuition for which instances get solved faster by \textit{Concorde} or by \textit{MSLS}. This is shown in Figure \ref{fig:landfeatures}, where the density (a), the diameter (b), the standard deviation of the degree distribution (c) and the skewness of the degree distribution (d) are shown. When we compare these figures to the difference in runtime of Figure \ref{fig:pcafill}, we see that graphs with a low density are likely to be solved faster by \textit{Concorde}. There is less correlation with the other features however. We do see an interesting region in the bottom left of the skewness plot in Figure \ref{fig:landfeatures} (d). These instances have very low skewness values and are in roughly the same region as the instances with very high differences in runtime (5+ seconds). A low skewness value is not a sufficient condition to be solved much faster by \textit{Concorde} however. The instance also needs to have a density of around 0.3, a diameter of 2, etc. This list of specific features gives us a good idea of what these instances look like. For example, the negative skewness of the degree distribution tells us that these are graphs where most nodes have a very large degree. 
	%Combining all these features results in a good prediction of whether or not an instance will have a very high difference in runtime, with a false negative rate of 0.2\% and a false positive rate of 2.4\% on our set of instances. 
	%Note that these are instances which \textit{MSLS} has a lot of trouble solving, with runtimes of up to 165 seconds, whereas \textit{Concorde} solves these in less than a second.
	Note that \textit{MSLS} has a lot of trouble solving these instances, with runtimes of up to 165 seconds, whereas \textit{Concorde} solves them in less than a second.  Moreover, \textit{MSLS} did not find the optimal solution in the large majority of these cases.
	
	We presented these results to the authors of \textit{MSLS}, in order to find a reason as to why their algorithm has such a hard time solving these graphs. Some hypotheses were proposed, such as the parameters of \textit{MSLS} being specifically tuned to solve graphs of several thousands of nodes. In that case we would expect the performance of \textit{MSLS} to be equally bad on all our instances, and not only in one very particular region.
	We tried using parameters that were more fit for smaller graphs and were able to reduce runtimes from around 130 seconds to around 12 seconds, but this is still much longer than other graphs. Perhaps these graphs posses certain structural properties which \textit{Concorde} handles in an optimized way, for example by using problem decomposition. The cause of these large runtimes is still unknown. 
	
\begin{figure*} 
		\centering
		\includegraphics[width=.75\textwidth,center]{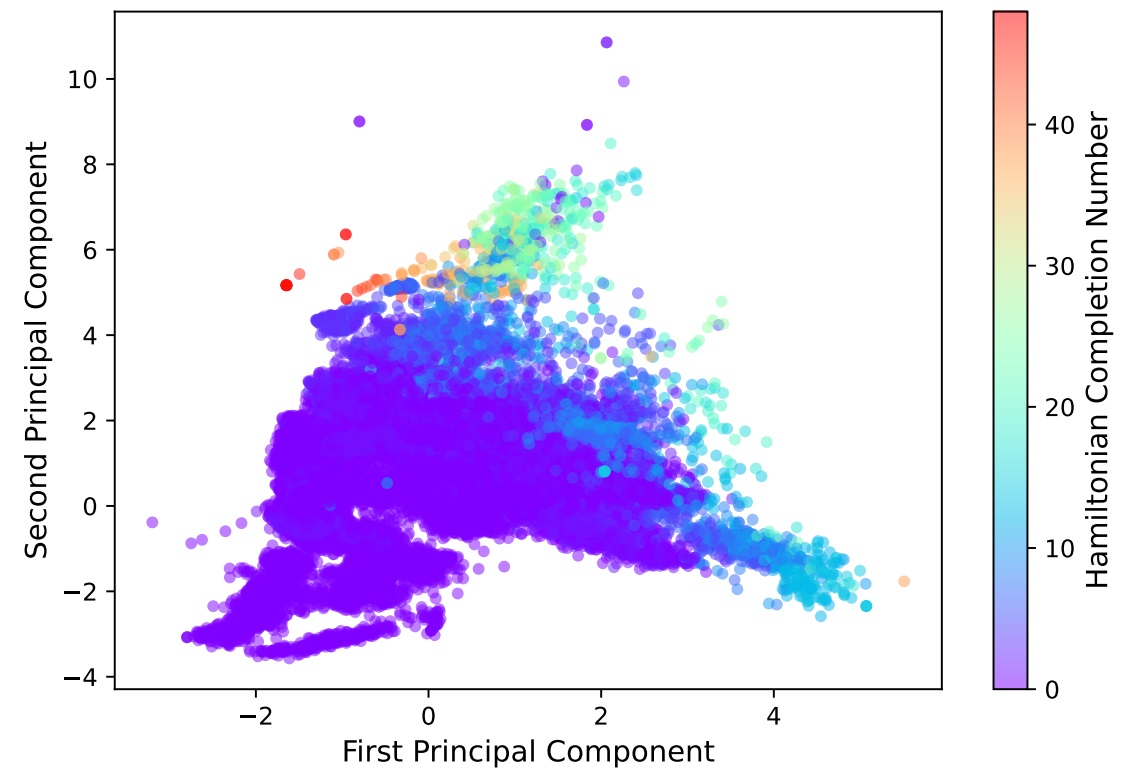}
		\caption{The Hamiltonian Completion number for each of the instances}
		\label{fig:hcp}
	\end{figure*}

	\textcolor{blue}{
	Finally we can also plot the Hamiltonian Completion number, which is the number of edges required to solve the completion problem. This is illustrated in Figure \ref{fig:hcp}, where instances with a high number are mostly (but not always) solved faster by \textit{Concorde} and vice versa. This can again give some intuitions about the algorithms, as this means that in most cases \textit{MSLS} can find the solution faster than \textit{Concorde} if the Hamiltonian Completion number is 0, so if there is already a Hamiltonian cycle.}
	
	%Each 1 column

	% 2 column fitting image

	\subsection{Classification}
\begin{figure*}[h]
		\centering
		\includegraphics[width=.7\textwidth,center]{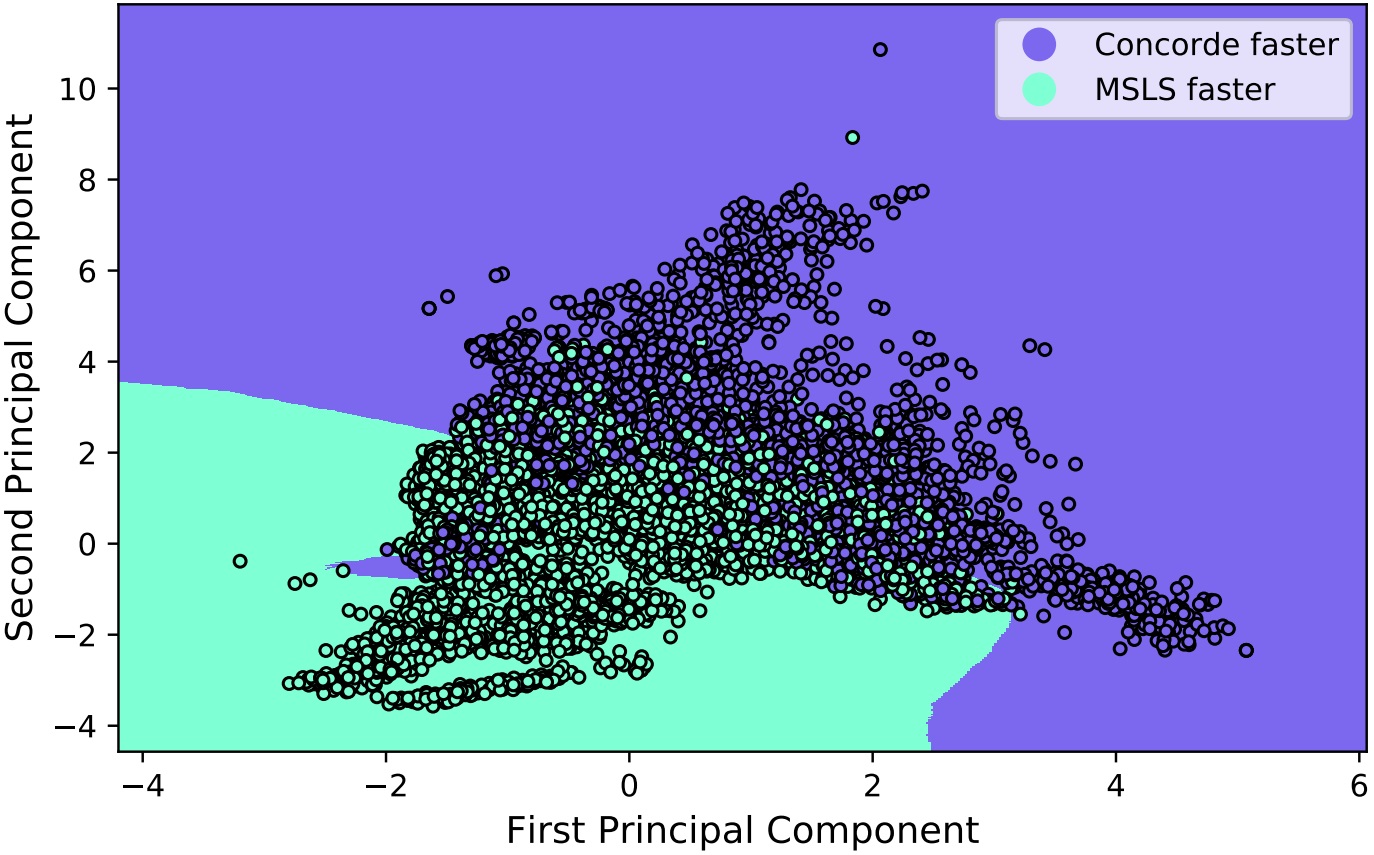}
		\caption{Binary k-nearest neighbors classification and the training data (k = 100)}
		\label{fig:nneighbors}
	\end{figure*}

	Now that we have analyzed the correlation of instance features with the difference in runtime, we can predict algorithm performance on graphs based on their location in the landscape. We first split the instances into a training and a testing set, each with 50\% of the instances. We then used a k-nearest neighbors classification algorithm on the training set, with k = 100. That resulted in every spot on the landscape being classified as either being faster with \textit{Concorde} or with \textit{MSLS}. This can be seen in Figure \ref{fig:nneighbors}, where the training data is also plotted. 
	With our nearest neighbors classification, 80.6\% of the test set was classified correctly. The large majority of the instances that were classified incorrectly were ones located around the border between instances that are solved faster by \textit{MSLS} and ones that are solved faster by \textit{Concorde}. The instances with a very large difference in runtime were also likely to be classified incorrectly.
	Overall, this classification is very promising. We projected instances onto two principal components, retaining only 69.0\% of the variance, but still managed to predict with good accuracy which algorithm would solve them faster. This suggests that our methodology is effective at reducing a complex instance space down to two dimensions, while still retaining key characteristics.

	% 2 column fitting image
	
	\section{Related work}
	\label{sec:related}
	Our methodology was largely based on similar research publications. The process of evolving instances that are difficult to solve for certain algorithms and then analyzing the properties of the generated instances was demonstrated by \cite{Evolvinginstances} on TSP instances.
	\textcolor{blue}{TSP has been analyzed using multiple different fitness functions, including the number of local search operations \citep{EvolvedInstancesTSP}, approximation quality \citep{approxquality,approxquality2} and multiple objectives \citep{multobjectives}. The performance of different heuristics have also been investigated, both with the goal of maximizing performance differences \citep{maxperformance,maxperformance2} and feature diversity \citep{maxdiversitybossek}. Besides TSP, the knapsack problem \citep{knapsack}, the quadratic knapsack problem \citep{quadknapsack}, and the graph coloring  problem \citep{SMITHMILES2015102} have also been investigated in this way. \textcolor{cyan}{Hard instances of the 0-1 knapsack problem are also studied in \cite{pisinger,jooken2,smithmilesKnap}}. Some recent papers have shifted their focus to increasing the diversity of the evolved instances, as only looking at performance can sometimes lead to sets of similar instances \citep{maxdiversity,featurediversity}. Examples include machine learning classification \citep{featurediversityML}, black box optimization \citep{featurediversityBB}, constrained subset selection \citep{constrained} and the minimum spanning tree problem \citep{minspantree}. Additionally, problem independent schemes have been proposed in order to generate instances that are diverse with respect to multiple features at the same time \citep{multifeatdiverse,multifeatdiverse2}. Recently \cite{morethan2} have shown that it is possible to generate instances where the performance of more than just two algorithms differs.} 
	
	The visualization and analysis of the instance space was based on numerous publications that are part of the Melbourne Algorithm \& Test Instance Library with Data Analytics (MATILDA). 
	This is a research platform developed via the project “Stress-testing algorithms: generating new test instances to elicit insights”, which aims to develop new methodologies and tools to reveal the strengths and weaknesses of algorithms. MATILDA focuses on the visualization of instance spaces, showing the strengths and weaknesses of algorithms, and analyzing instances that algorithms find easy or hard. Their methodology was first described in three papers focusing on graph coloring, namely \cite{smith1}, \cite{smith2} and \cite{SMITHMILES2015102}. These methods have also been applied on several other problem domains, including combinatorial optimization problems, continuous black-box optimization, machine learning classification and time series forecasting. The full set of publications can be found on the official website of MATILDA \citep{matilda}.

	\section{Conclusion}
	\label{sec:conclusion}
	Generating or selecting a solid set of test instances is key when comparing algorithms. However, randomly generating instances or using standard graph generators can result in a lack of diversity and difficulty. Using established benchmark instances can also play a part in the over-tuning of algorithms.
	For these reasons we used a novel methodology which utilizes evolutionary algorithms to generate diverse and challenging instances. We applied this method to the Hamiltonian Completion Problem in order to compare the performance of \textit{Concorde} to the performance of a multi-start local search algorithm. We were able to generate a very diverse set of instances by first evolving in instance hardness and then filling in the landscape, which allowed us to gain new insights into both algorithms. Visualizing the instance space provided a better understanding of their strengths and weaknesses and showed that standard graph generators have a limited footprint. It also demonstrated that reducing dimensions retains key characteristics, including the ability to predict algorithm performance based on the location of an instance in the landscape. 
	%The last sections displayed the power of visualization, offering several intuitions on the topics of evolutionary algorithms, algorithm performance and instance features.
	In summary, our work has provided an instance space analysis of a combinatorial optimization problem, further supporting this methodology. \textcolor{blue}{This can help researchers to develop better solvers, further optimize existing ones, and help in benchmarking.}
	
	Future research could experiment with another solver, for example using Lin-Kernighan-Helsgaun and analyzing its performance in the landscape. \textcolor{blue}{Using different evolutionary operators can also be interesting.} Another interesting idea that was not explored in this paper is the extrapolation of our newly gained insights to larger graphs. 
	For example, whether or not graphs with a few thousand nodes can be projected onto the same landscape and if our current classifier would be able to predict algorithm performance with reasonable accuracy.

%	\section*{CRediT authorship contribution statement}
%	\textbf{Thibault Lechien:} Conceptualization, Methodology, Software, Validation, Formal analysis, Data Curation, Writing - Original Draft, Writing - Review \& Editing, Visualization.
%	\textbf{Jorik Jooken:} Conceptualization, Methodology, Software, Writing - Review \& Editing.
%	\textbf{Patrick De Causmaecker:} Conceptualization, Methodology, Supervision, Project administration.
	
	\section*{Conflict of Interest Statement}
	Declarations of interest: none. 
	
	\section*{Acknowledgements}
	We gratefully acknowledge the support provided by the ORDinL project (FWO-SBO S007318N, Data Driven Logistics,
	1/1/2018 - 31/12/2021). This research received funding from the Flemish Government under the “Onderzoeksprogramma
	Artificiële Intelligentie (AI) Vlaanderen” programme.

	\bibliographystyle{elsarticle-harv}
	\bibliography{bibliography}

	\appendix
	\renewcommand\thefigure{\thesection.\arabic{figure}}  
	\section{Supplementary figures}
	\setcounter{figure}{0}    
	\begin{figure*}[h]
		\centering
		%3x2 single column images
		\begin{subfigure}{.49\textwidth}
			\centering
			\includegraphics[width=.99\textwidth,center]{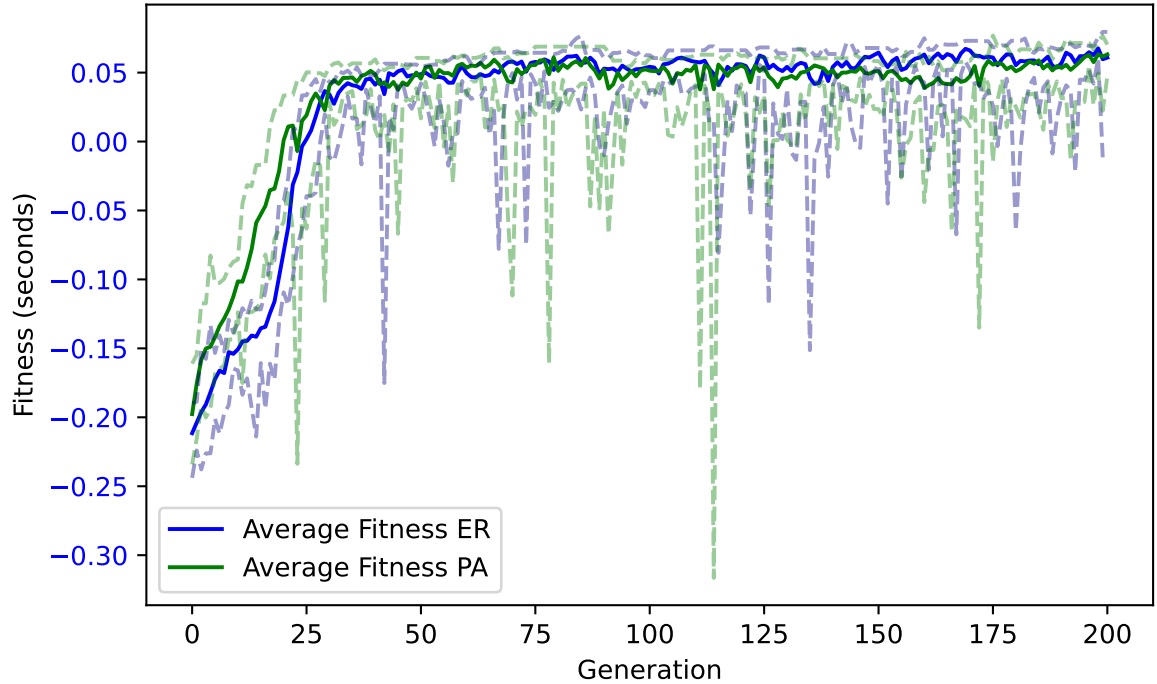}
			\caption{The fitness of the population}
		\end{subfigure}
		\begin{subfigure}{.49\textwidth}
			\centering
			\includegraphics[width=.97\textwidth,center]{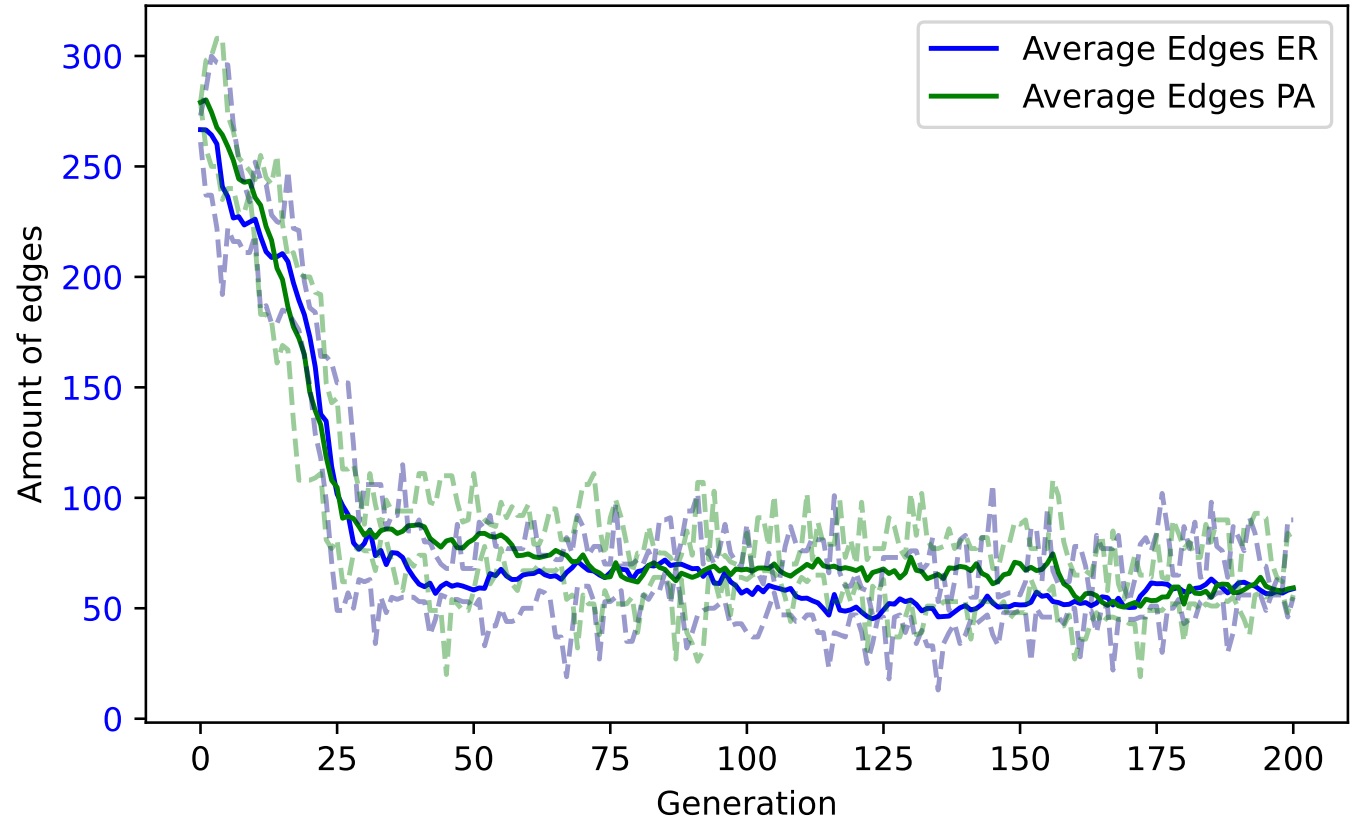}
			\caption{The number of edges of the population}
		\end{subfigure}
		\begin{subfigure}{.49\textwidth}
			\centering
			\includegraphics[width=.99\textwidth,center]{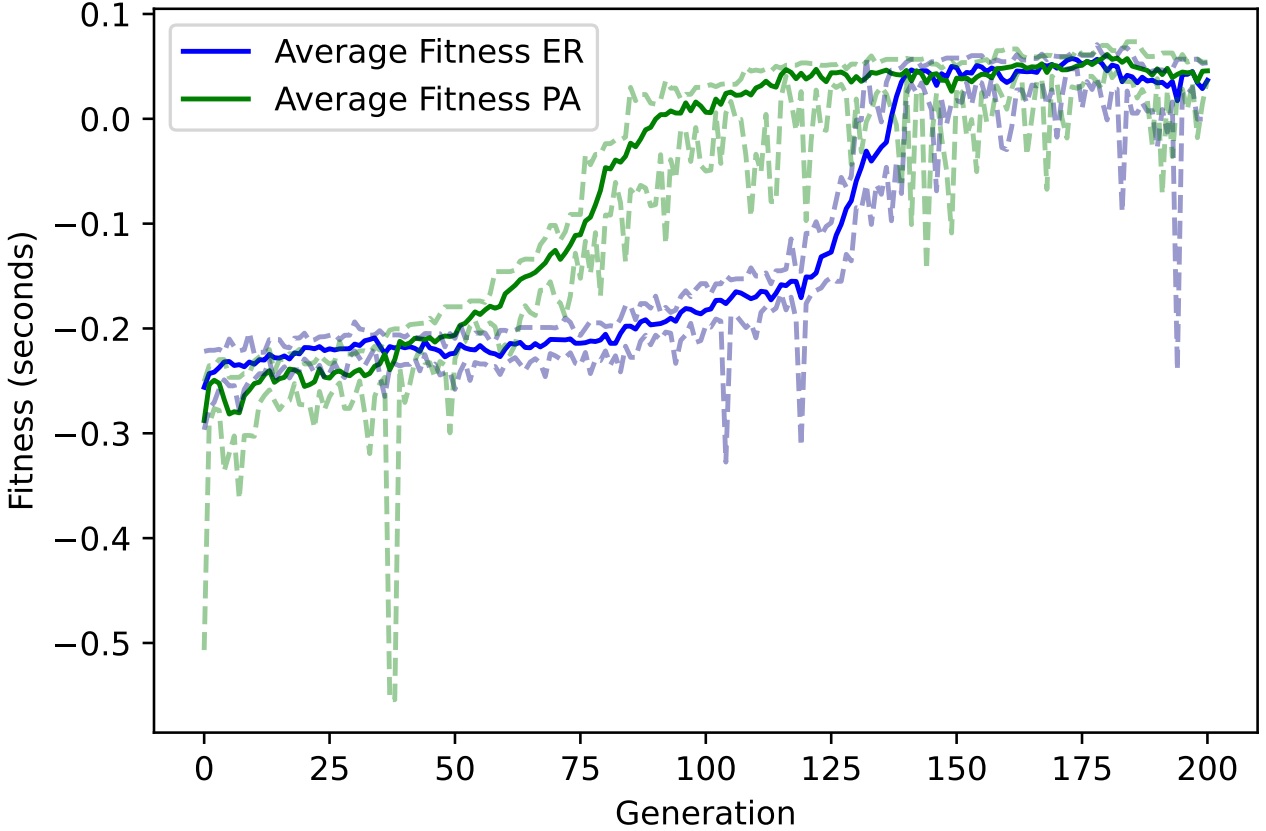}
			\caption{The fitness of the population}
		\end{subfigure}
		\begin{subfigure}{.49\textwidth}
			\centering
			\includegraphics[width=.97\textwidth,center]{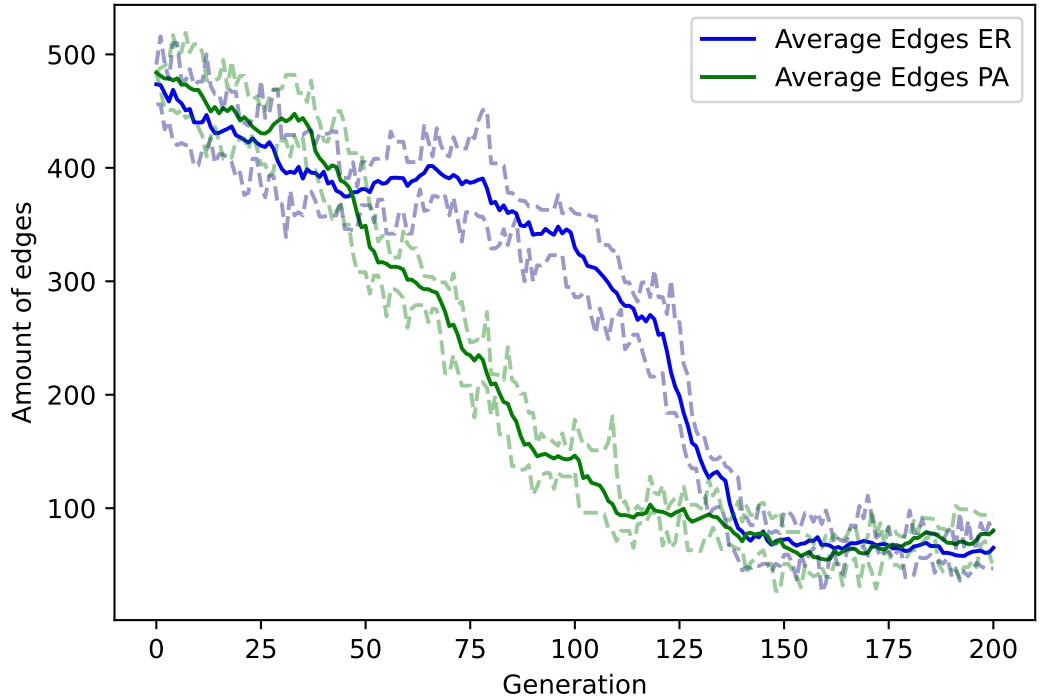}
			\caption{The number of edges of the population}
		\end{subfigure}
		\begin{subfigure}{.49\textwidth}
			\centering
			\includegraphics[width=.99\textwidth,center]{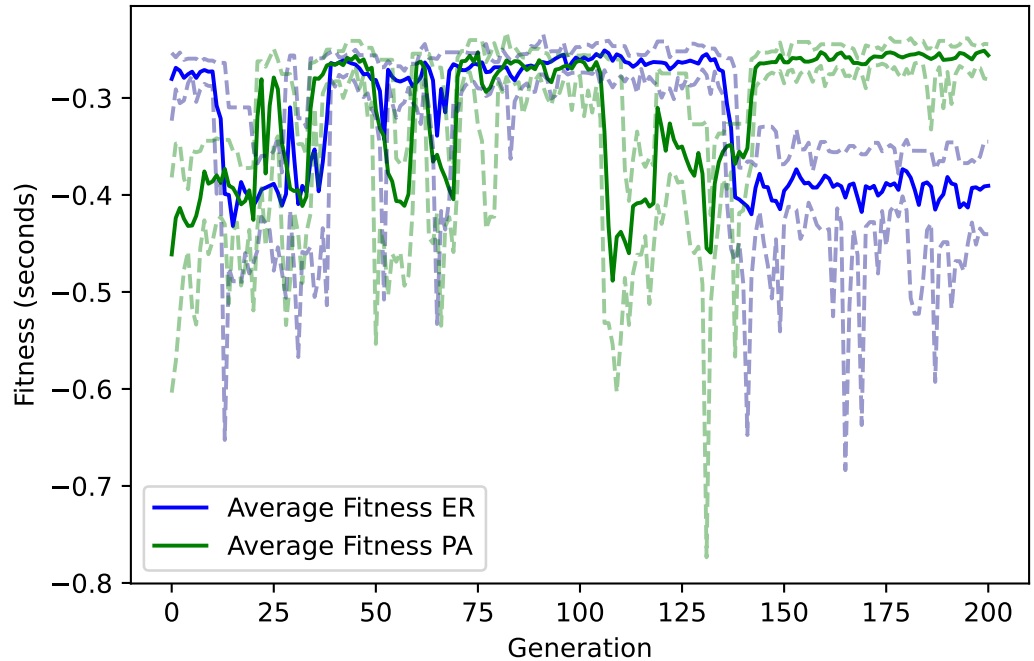}
			\caption{The fitness of the population}
		\end{subfigure}
		\begin{subfigure}{.49\textwidth}
			\centering
			\includegraphics[width=.97\textwidth,center]{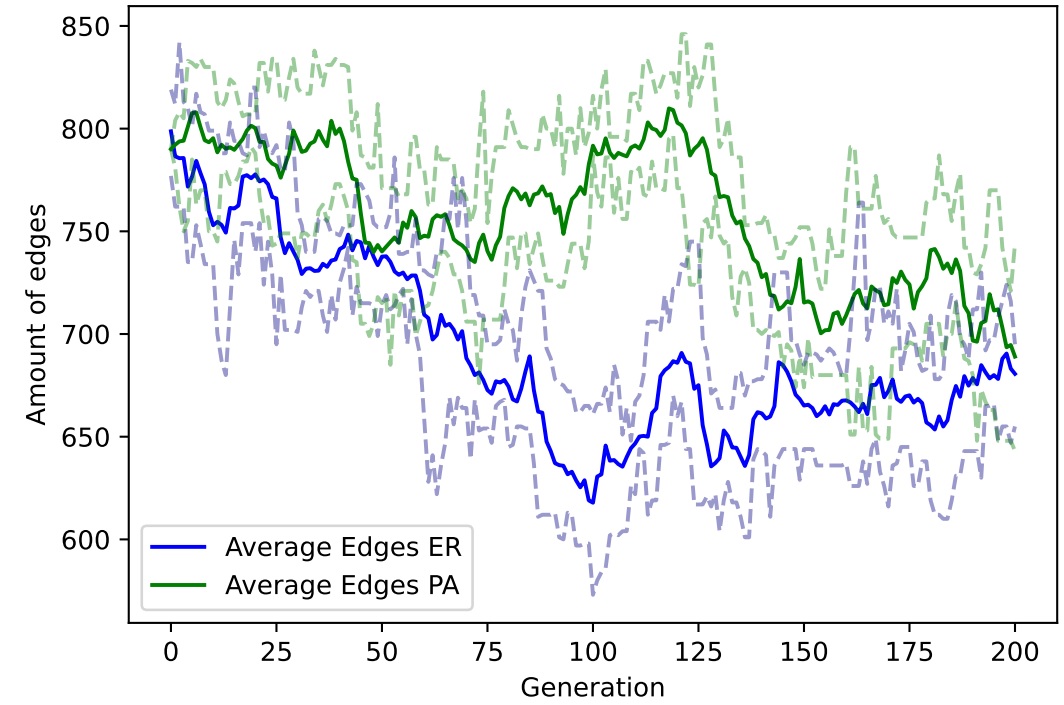}
			\caption{The number of edges of the population}
		\end{subfigure}
		\caption{\textcolor{cyan}{Statistics of sample runs of the evolutionary algorithm that maximizes fitness with the goal of analyzing the difference that the starting population makes. Each row represents different runs, using different initial edge counts (300, 500 and 800 respectively). Blue lines indicate a starting population of Erdős-Rényi graphs (ER) and the green lines indicate a starting population of preferential attachment graphs (PA).}}
		\label{fig:EvoPA}
	\end{figure*}

	\begin{figure*}[h]
		\centering
		%Both single column images
		\begin{subfigure}{.49\textwidth}
			\centering
			\includegraphics[width=.99\textwidth,center]{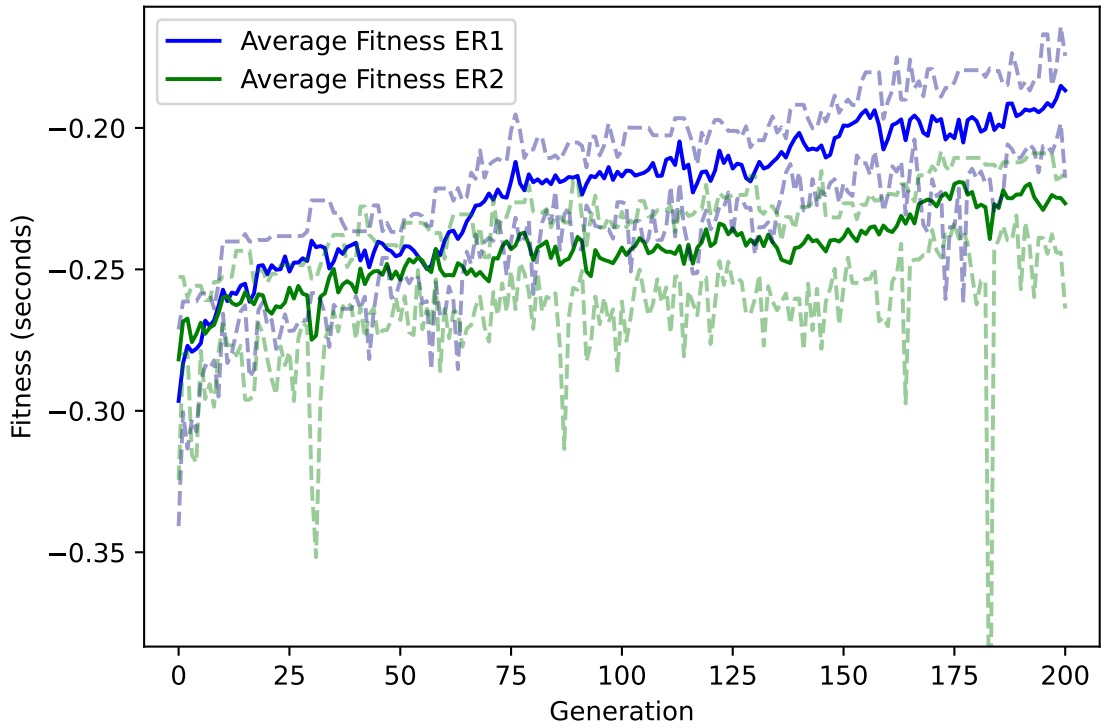}
			\caption{The fitness of the population}
		\end{subfigure}
		\begin{subfigure}{.49\textwidth}
			\centering
			\includegraphics[width=\textwidth,center]{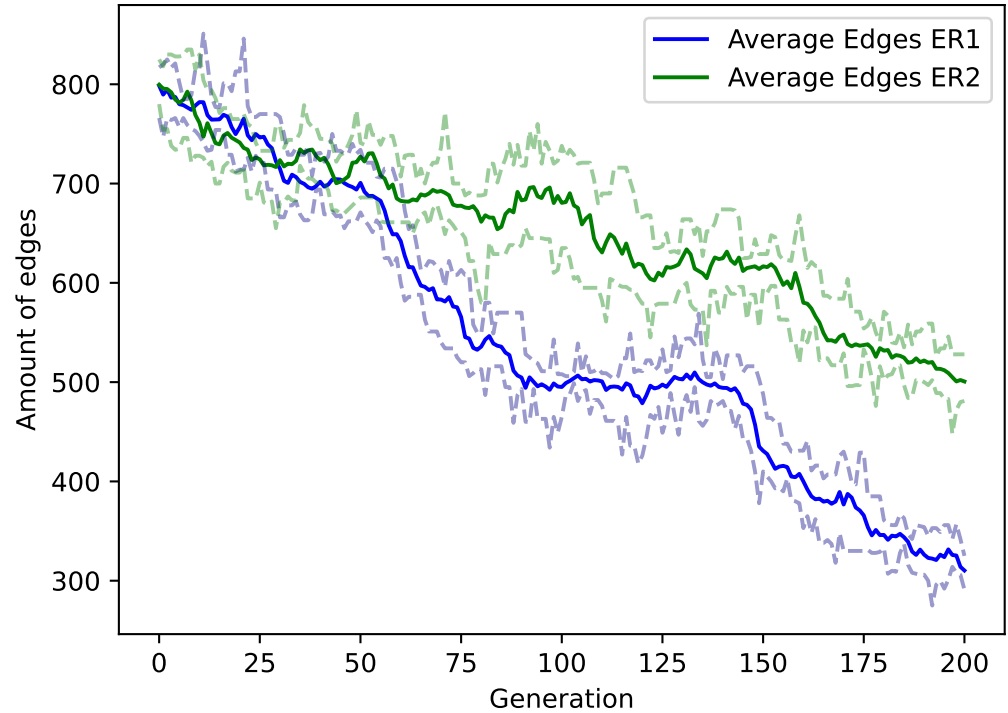}
			\caption{The number of edges of the population}
		\end{subfigure}
		\caption{\textcolor{cyan}{Statistics of sample runs of the evolutionary algorithm that maximizes fitness with the goal of analyzing the difference that the starting population makes. The algorithm was ran twice, so the difference between the first and second run is the (randomly generated) starting population.}}
		\label{fig:EvoER}
	\end{figure*}

	\begin{figure*}[h]
		%Double column image
		\centering
		\includegraphics[width=.7\textwidth,center]{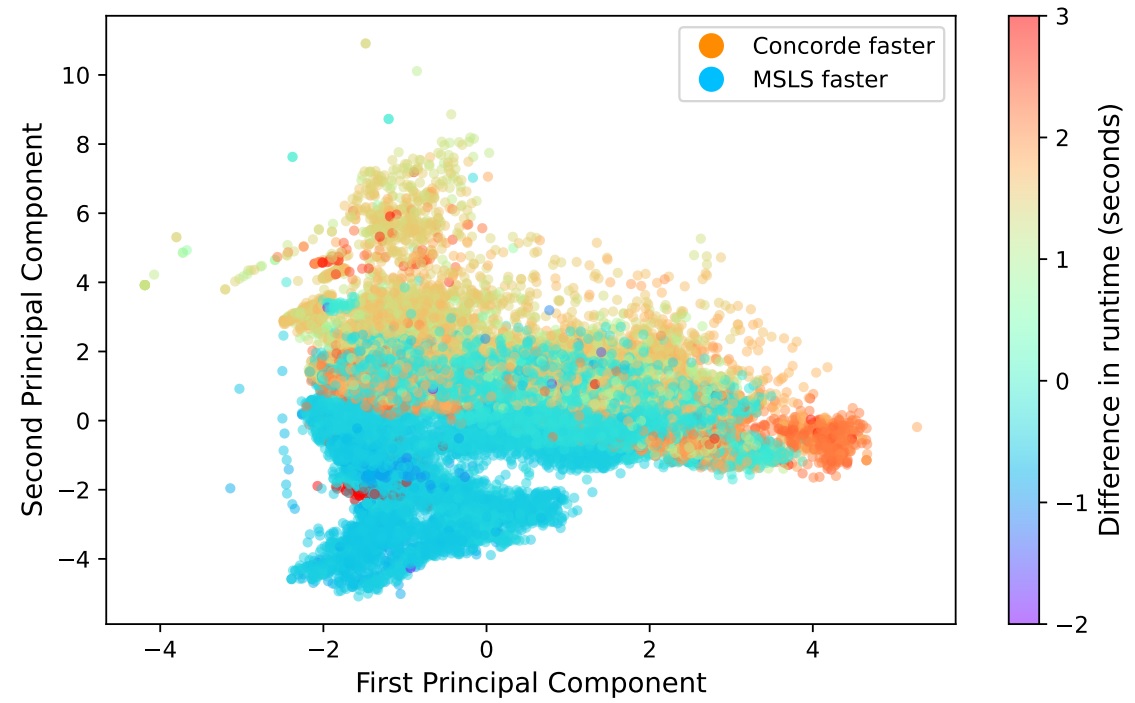}
		\caption{\textcolor{cyan}{The result of fitting the principal component analysis on the full set of instances instead of on the initial set.}}
		\label{fig:pca_again}
	\end{figure*}

%	\begin{thebibliography}{00}
%		
%		%% \bibitem{label}
%		%% Text of bibliographic item
%		
%		\bibitem{}
%		
%	\end{thebibliography}

\end{document}